\documentclass[sigconf, screen,nonacm]{acmart}
\AtBeginDocument{%
  }
\usepackage{multirow}  
\usepackage[most]{tcolorbox}
\usepackage{multicol}
\definecolor{Gray}{gray}{0.9}
\definecolor{myblue}{RGB}{230, 230, 248}
\usepackage{stfloats}
\setcopyright{acmlicensed}
\copyrightyear{2018}
\acmYear{2018}
\acmDOI{XXXXXXX.XXXXXXX}
\acmConference[Conference acronym 'XX]{Make sure to enter the correct
  conference title from your rights confirmation email}{June 03--05,
  2018}{Woodstock, NY}

\begin{document}

\title{PianoFlow: Music-Aware Streaming Piano Motion Generation with Bimanual Coordination}

\author{Xuan Wang}
\affiliation{%
  \institution{Zhejiang University}
  \city{Hangzhou}
  \state{Zhejiang}
  \country{China}
}

\author{Kai Ruan}
\affiliation{%
  \institution{Renmin University of China}
  \city{Beijing}
  \country{China}}

\author{Jiayi Han}
\affiliation{%
  \institution{Beijing Forestry University}
  \city{Beijing}
  \country{China}}

\author{Kaiyue Zhou}
\affiliation{%
  \institution{Chengdu Minto Tech}
  \city{Chengdu}
  \state{Sichuan}
  \country{China}}

\author{Gaoang Wang}
\authornote{Corresponding author.}
\affiliation{%
  \institution{Zhejiang University}
  \city{Hangzhou}
  \state{Zhejiang}
  \country{China}}

\renewcommand{\shortauthors}{Wang et al.}

\begin{abstract}
Audio-driven bimanual piano motion generation requires precise modeling of complex musical structures and dynamic cross-hand coordination. However, existing methods often rely on acoustic-only representations lacking symbolic priors, employ inflexible interaction mechanisms, and are limited to computationally expensive short-sequence generation. To address these limitations, we propose PianoFlow, a flow-matching framework for precise and coordinated bimanual piano motion synthesis. Our approach strategically leverages MIDI as a privileged modality during training, distilling these structured musical priors to achieve deep semantic understanding while maintaining audio-only inference. Furthermore, we introduce an asymmetric role-gated interaction module to explicitly capture dynamic cross-hand coordination through role-aware attention and temporal gating. To enable real-time streaming generation for arbitrarily long sequences, we design an autoregressive flow continuation scheme that ensures seamless cross-chunk temporal coherence. Extensive experiments on the PianoMotion10M dataset demonstrate that PianoFlow achieves superior quantitative and qualitative performance, while accelerating inference by over 9× compared to previous methods.

\end{abstract}



\maketitle
\section{Introduction}

\begin{figure}
    \centering
    \includegraphics[width=\linewidth]{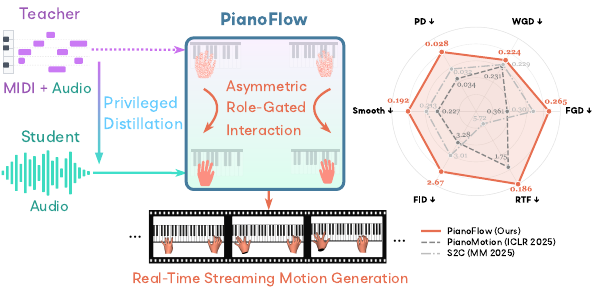}
    \caption{PianoFlow enables high-fidelity and real-time bimanual piano motion generation. (Left) Our audio-driven framework distills privileged musical priors and employs Asymmetric Role-Gated Interaction for coordinated bimanual synthesis. (Right) PianoFlow achieves state-of-the-art performance across all metrics, supporting seamless real-time generation.}
    \label{fig:teaser}
\end{figure}

Audio-driven human motion generation has progressed significantly, enabling applications in VR/AR, digital humans, and robotics~\cite{zhu2023human,sui2026survey,jiang2024audio}. This field encompasses well-studied subproblems, such as audio-driven dance~\cite{siyao2022bailando,li2024lodge,tseng2023edge}, talking head~\cite{shen2023difftalk,peng2023emotalk,richard2021meshtalk}, and co-speech gesture generation~\cite{zhi2023livelyspeaker}. Recently, research has expanded to instrument performance synthesis, exploring tasks such as violin~\cite{nishizawa2025syncviolinist,jin2024audio}, Guzheng~\cite{chen2021music}, and piano~\cite{pm} motion generation. Unlike conventional tasks that rely on coarse rhythmic or phonetic alignment, instrument performance demands precise, high-frequency finger articulation, strict physical interaction constraints, and deep alignment with structured musical harmony. In this work, we focus on the challenging task of audio-driven bimanual piano motion generation, tackling the unique challenge of dynamic cross-hand coordination across a dense spatial keyboard.

Early works like PianoPlayer~\cite{musy_pianoplayer} generate static hand poses from piano scores, relying on classifiers to predict fingering. Subsequently, BACH~\cite{jiao2025bach} transforms scores into continuous hand motion using a two-stage deep generative framework that predicts keyframe guidance and synthesizes smooth transitions. Driven by the emergence of aligned datasets, recent research has shifted toward audio-conditioned generation. PianoMotion~\cite{pm} leverages diffusion models to generate continuous sequences from audio, while S2C~\cite{s2c} decouples hands and models them separately, followed by a static interaction mechanism.

However, synthesizing realistic bimanual motion directly from continuous audio remains challenging due to three primary limitations in existing methods:
(1) \textbf{Limitations of acoustic-only representations:} Existing methods often rely on speech-based encoders that lack music-specific sensitivity. Furthermore, without explicit symbolic priors, they struggle to capture fine-grained note onsets and harmonic structures from complex audio.
(2) \textbf{Inflexible bimanual coordination:} The hands assume distinct, time-varying roles (e.g., melody vs.\ accompaniment). Static rules or weak coupling fail to capture these dynamic shifts, often resulting in bimanual interference.
(3) \textbf{Limited scalability and real-time efficiency:} Most frameworks are computationally intensive and restricted to fixed-length sequences, precluding streaming deployment for arbitrarily long performances.

To address these challenges, we propose \textbf{PianoFlow}, a two-stage flow-matching framework as illustrated in Figure~\ref{fig:teaser}. The system comprises Music-Aware Wrist Trajectory Generation and Bimanual Coordinated Gesture Generation. Unlike prior works using speech encoders like Wav2Vec2.0~\cite{baevski2020wav2vec} or Hubert~\cite{hsu2021hubert}, we adopt MuQ~\cite{zhu2025muq} for music-specific representations. Furthermore, we leverage MIDI~\cite{rothstein1995midi} as a privileged modality, employing a Harmonic Perceiver to capture octave and interval relationships. By distilling a multimodal teacher into an audio-only student, we inject MIDI-level structural priors into the learned representations while maintaining audio-driven inference~\cite{mansourian2025comprehensive,aslam2023privileged}.
Additionally, we adopt a decoupled bimanual formulation while enabling interaction at the feature bottleneck. We propose an Asymmetric Role-Gated Interaction module (ARGI), which utilizes role-specific biases as identity cues to facilitate information exchange. A temporal gating mechanism further modulates interaction strength, enabling adaptive coordination while reducing interference during role divergence.

Finally, to support real-time streaming for long sequences, we introduce Autoregressive Flow Continuation (AFC). Unlike conventional methods that suffer from boundary discontinuities or require heavy context-caching~\cite{li2024lodge,siyao2022bailando,fan2022faceformer}, AFC operates directly during the ODE integration phase. It maintains temporal consistency by anchoring overlapping regions to the historical trajectory, ensuring seamless transitions between adjacent chunks. This design allows PianoFlow to achieve coherent long-form generation with a Real-Time Factor (RTF) of 0.186, providing a speedup of over 9$\times$ compared to prior methods.

Extensive experiments on PianoMotion10M, encompassing both short- and long-sequence generation, demonstrate the superiority of our architecture through quantitative and qualitative comparisons. Furthermore, ablation studies investigate the specific role of each individual component. These results confirm that PianoFlow effectively interprets musical semantics to synthesize precise and coordinated playing motions.

The main contributions of this work are summarized as follows:
\begin{itemize}
    \item We propose \textbf{PianoFlow}, a two-stage framework for audio-driven bimanual motion generation. By leveraging MIDI as a privileged modality, it distills structural priors to achieve a deep understanding of musical semantics while maintaining audio-driven inference.
    
    \item We design an Asymmetric Role-Gated Interaction mechanism to explicitly model bimanual dynamics. By incorporating role-specific biases and temporally modulating interaction strength, this design enables fine-grained and adaptive coordination between the two hands.
    
    \item We develop Autoregressive Flow Continuation, ensuring cross-chunk temporal coherence and real-time, arbitrary-length streaming inference by anchoring to the optimal transport path.
    
    \item Extensive experiments on the PianoMotion10M dataset demonstrate that PianoFlow achieves superior performance in both quantitative and qualitative evaluations, while accelerating inference by over $9\times$ compared to prior methods.
\end{itemize}

\section{Related Work}

\subsection{Audio-Driven Motion Generation}

Recent advances in generative modeling have significantly improved audio-driven human motion synthesis across diverse applications, including co-speech motion generation~\cite{shen2023difftalk,sun2024diffposetalk,aneja2024facetalk,cassell2001beat,zhi2023livelyspeaker,liu2024emage,yi2023generating,chu2025artalk,jafari2024jambatalk,liu2024towards,xu2024mambatalk,chhatre2024emotional,chen2025language,fan2024unitalker,kim2025deeptalk,xu2026emotionally,liu2022disco,chae2025perceptually,kim2025memorytalker,liu2025semges}, music-driven dance generation~\cite{siyao2022bailando,tseng2023edge,li2023finedance,yang2025matchdance,li2024lodge,guo2025controllable,dai2026tcdif,liu2024dgfm,liu2025gcdance,ghosh2025duetgen,yang2025megadance,li2025music,wang2025pamd,fan2025align,huang2024beat}, and specialized scenarios~\cite{xu2025mospa}. Early methods focused on learning direct mappings from audio to motion using adversarial or variational frameworks~\cite{kim2022brand}. In contrast, recent approaches increasingly leverage large-scale generative architectures, such as Transformers and diffusion models, to capture complex spatio-temporal dynamics~\cite{sun2024diffposetalk}.
To enhance motion realism and temporal coherence, many strategies have been proposed. For instance, several pioneering works employ explicit structural representations to achieve accurate synchronization and expressive generation~\cite{richard2021meshtalk,zhang2023sadtalker}. Transformer-based models have further enhanced the modeling of long-range temporal dependencies~\cite{fan2022faceformer}, while diffusion-based approaches have demonstrated the ability to synthesize expressive motion sequences directly from audio without requiring heavy intermediate structural constraints~\cite{wei2024aniportrait,liu2025semges}.
However, synthesizing realistic piano motions requires parsing complex harmonic structures and fine-grained note onsets, a rigorous challenge absent in general body dynamics. The reliance of existing models on coarse-grained acoustic features limits their ability to meet the stringent demands of precise finger articulation. To bridge this semantic gap, we introduce a cross-modal distillation scheme to embed explicit musical priors into the audio representation, establishing a highly precise semantic foundation tailored for piano performance.

\subsection{Piano Hand Motion Generation}

Piano hand motion generation requires the precise modeling of fine-grained finger articulation and coordinated bimanual interaction. Existing approaches generally follow two paradigms: embodied robotic control and data-driven kinematic synthesis. 
The former focuses on learning dexterous manipulation within physical simulators via reinforcement or imitation learning~\cite{zakka2023robopianist,zeulner2025learning}. For instance, RoboPianist~\cite{zakka2023robopianist} established a physics-based framework leveraging deep RL and fingering priors for complex bimanual performances, while FürElise~\cite{wang2024furelise} integrated generative modeling by guiding a physics-based controller with diffusion-generated kinematic trajectories. 
The latter paradigm synthesizes hand kinematics directly from musical signals, supporting expressive applications such as AI-assisted education and digital human animation. Within this data-driven setting, Gan \textit{et al.}~\cite{pm} introduced the PianoMotion10M dataset alongside a diffusion-based baseline, while Liu \textit{et al.}~\cite{s2c} modeled bimanual coordination through a dual-stream diffusion architecture. However, these methods rely on static interactions that fail to capture dynamic, asymmetric cross-hand role shifts (e.g., melody vs. accompaniment). Furthermore, their diffusion pipelines suffer from prohibitive sampling latency and are restricted to short-sequence generation. To overcome these bottlenecks, PianoFlow introduces an Asymmetric Role-Gated Interaction (ARGI) module to explicitly model time-varying bimanual dependencies, alongside an Autoregressive Flow Continuation (AFC) mechanism for seamless, real-time streaming synthesis of arbitrary-length sequences.

\begin{figure*}
    \centering
    \includegraphics[width=\textwidth]{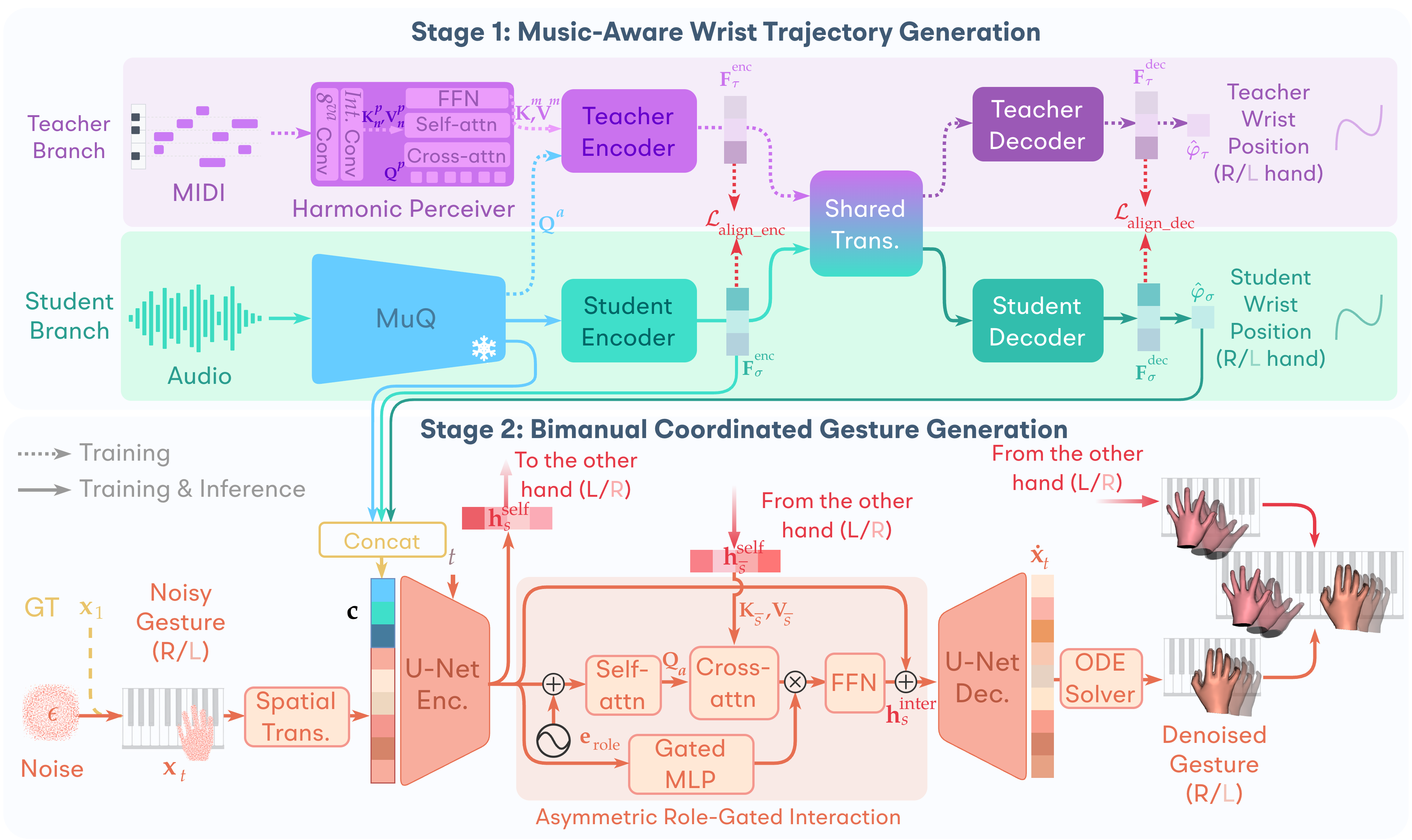}
    \caption{Overview of the PianoFlow architecture. The framework operates in two stages: (1) Music-Aware Wrist Trajectory Generation utilizes a decoupled cross-modal distillation scheme to transfer structured harmonic knowledge from a multimodal teacher (MIDI and audio) to an audio-only student. (2) Bimanual Coordinated Gesture Generation synthesizes fine-grained kinematics via conditional flow-matching. Specifically, a factorized spatio-temporal backbone estimates the velocity field, while an Asymmetric Role-Gated Interaction (ARGI) module explicitly models bimanual coordination at the network bottleneck.}
    \label{fig:arch}
\end{figure*}
\section{Method}
\label{sec:method}

Our goal is to synthesize precise and well-coordinated bimanual piano motions directly from audio. As illustrated in Fig.~\ref{fig:arch}, the proposed PianoFlow framework consists of two stages: (1) Music-Aware Wrist Trajectory Generation and (2) Bimanual Coordinated Gesture Generation. In the first stage, we introduce a cross-modal distillation scheme that leverages MIDI as a privileged modality, transferring structured musical priors to the audio-driven student model. In the second stage, we employ a spatio-temporal flow-matching framework to model fine-grained finger kinematics, incorporating an Asymmetric Role-Gated Interaction (ARGI) mechanism to explicitly capture dynamic cross-hand coordination. For real-time inference, we adopt an Autoregressive Flow Continuation (AFC) strategy, enabling coherent streaming synthesis for sequences of arbitrary length.

\subsection{Music-Aware Wrist Trajectory Generation}
\label{sec:stage1}

Given an input audio sequence, this stage predicts bimanual wrist trajectories $\varphi=[\varphi^L, \varphi^R] \in \mathbb{R}^{N \times 6}$, where $N$ denotes the sequence length. To explicitly ground these trajectories in musical semantics, a cross-modal distillation scheme transfers structured harmonic priors from a multimodal (MIDI and audio) teacher to an audio-only student. Considering the distinct behaviors of each hand, we model their trajectories using independent networks with identical architectures but unshared parameters. This decoupled design prevents premature coupling, allowing each branch to capture hand-specific kinematic priors.

\subsubsection{Decoupled Cross-Modal Architecture.}

We adopt a decoupled teacher-student architecture, where modality-specific features are learned independently and aligned in a shared latent space.

\noindent\textbf{Student Branch Encoding.}
Taking raw audio as input, we extract acoustic features using a frozen MuQ~\cite{zhu2025muq} encoder, which is pre-trained on large-scale music datasets for robust music representation. These features are then processed through a Student Encoder to yield $\mathbf{F}_{\sigma}^{\text{enc}}$.

\noindent\textbf{Teacher Branch Encoding.}
The teacher branch leverages MIDI as a privileged modality to provide explicit symbolic priors. We design a Harmonic Perceiver to encode pitch-wise dependencies from the piano-roll input $\mathcal{X}_{\text{midi}} \in \mathbb{R}^{N \times 88}$, where each frame represents note activations across the 88 standard piano keys.

To capture the hierarchical nature of musical intervals within each frame $n$, we model the local harmonic structure along the pitch dimension using a cascaded convolutional design: an octave-aware kernel (size $13$) followed by an interval-aware kernel (size $5$). To retain the spatial keyboard topology, we explicitly inject an absolute pitch positional encoding $\mathbf{P}_{\text{pitch}} \in \mathbb{R}^{88 \times C}$. This yields a pitch-wise structural representation:
\begin{equation}
\mathbf{S}_n = \psi_{\text{int}}\left(\psi_{\text{oct}}(\mathcal{X}_{\text{midi}, n})\right) + \mathbf{P}_{\text{pitch}},
\end{equation}
where $\mathbf{S}_n \in \mathbb{R}^{88 \times C}$.

To aggregate this harmonic information, inspired by the asymmetric attention mechanism of the Perceiver \cite{Jaegle2021Perceiver}, we introduce a set of learnable queries to perform attention-based pooling over the pitch dimension:
\begin{equation}
\mathbf{Q}^{p} = \mathbf{Q}_0 \mathbf{W}_Q^{p}, \quad
\mathbf{K}_n^{p} = \mathbf{S}_n \mathbf{W}_K^{p}, \quad
\mathbf{V}_n^{p} = \mathbf{S}_n \mathbf{W}_V^{p},
\end{equation}
\begin{equation}
\tilde{\mathbf{E}}_n = \text{Softmax}\left(\frac{\mathbf{Q}^{p}(\mathbf{K}_n^{p})^\top}{\sqrt{d_p}}\right)\mathbf{V}_n^{p},
\end{equation}
where $\mathbf{Q}_0 \in \mathbb{R}^{M \times C}$ represents a set of $M$ learnable queries, $\mathbf{W}_Q^{p}, \mathbf{W}_K^{p}, \\ \mathbf{W}_V^{p}$ are learnable projection matrices, and $d_p$ denotes the hidden dimension of the pitch keys. 
This attention mechanism enables the model to selectively focus on distinct harmonic regions (e.g., melody and bass). Subsequently, we map $\tilde{\mathbf{E}}_n$ into a unified embedding $\mathbf{E}_n$. Finally, a temporal self-attention encoder processes the sequence $\{\mathbf{E}_n\}_{n=1}^N$ to yield the final MIDI representation $\mathbf{F}_{\text{midi}}$.

To fuse these modalities, the Teacher Encoder conditions the acoustic representation $\mathbf{F}_{\sigma}^{\text{enc}}$ on the aggregated MIDI features via cross-attention. 
Specifically, the acoustic features act as queries, while the MIDI features serve as keys and values:
\begin{equation}
\mathbf{Q}^{a} = \mathbf{F}_{\sigma}^{\text{enc}} \mathbf{W}_Q^{a}, \quad
\mathbf{K}^{m} = \mathbf{F}_{\text{midi}} \mathbf{W}_K^{m}, \quad
\mathbf{V}^{m} = \mathbf{F}_{\text{midi}} \mathbf{W}_V^{m},
\end{equation}
\begin{equation}
\tilde{\mathbf{F}} = \text{Softmax}\left(\frac{\mathbf{Q}^{a}(\mathbf{K}^{m})^\top}{\sqrt{d_m}}\right)\mathbf{V}^{m},
\end{equation}
\begin{equation}
\mathbf{F}_{\tau}^{\text{enc}} = \mathbf{F}_{\sigma}^{\text{enc}} + \tilde{\mathbf{F}},
\end{equation}
where $\mathbf{W}_Q^{a}, \mathbf{W}_K^{m}, \mathbf{W}_V^{m}$ are learnable projection matrices mapping the features to a joint attention space of dimension $d_m$. 
This operation explicitly injects structured musical priors into the acoustic representation, yielding the final teacher encoding $\mathbf{F}_{\tau}^{\text{enc}}$.

\noindent\textbf{Shared Backbone and Decoupled Decoding.}
We process the encoded student and teacher features, $\mathbf{F}_{\sigma}^{\text{enc}}$ and $\mathbf{F}_{\tau}^{\text{enc}}$, through a Transformer backbone shared across the student and teacher branches to map them into a unified latent space. Two decoupled decoding heads then generate $\mathbf{F}_{\sigma}^{\text{dec}}$ and $\mathbf{F}_{\tau}^{\text{dec}}$, respectively. A linear projection further maps each decoded feature to wrist trajectories, yielding $\varphi_{\sigma}, \varphi_{\tau} \in \mathbb{R}^{N \times 6}$. This design explicitly aligns the high-level representations within the shared space while preventing the student from directly fitting the teacher's exact output distribution.

\subsubsection{Cross-Modal Distillation.}

We achieve cross-modal distillation by aligning features at both the encoder and decoder stages. To ensure temporal consistency, we formulate the alignment loss by minimizing the cosine distance across the sequence dimension:
\begin{equation}
\mathcal{L}_{\text{align}} = \sum_{l \in \{\text{enc}, \text{dec}\}} \frac{1}{N} \sum_{n=1}^N 
\left( 1 - \frac{\langle \mathbf{F}_{\sigma, n}^l, \text{sg}(\mathbf{F}_{\tau, n}^l) \rangle}{\|\mathbf{F}_{\sigma, n}^l\|_2 \, \|\text{sg}(\mathbf{F}_{\tau, n}^l)\|_2} \right),
\end{equation}
where $\text{sg}(\cdot)$ denotes the stop-gradient operator.

While the MuQ acoustic encoder is pre-trained, the Harmonic Perceiver is trained from scratch. Consequently, enforcing strong alignment during early training may introduce unreliable supervision from the under-trained teacher. To mitigate this, we progressively increase the distillation loss weight $\lambda_{\text{distill}}$ during training. This strategy enables the model to learn robust modality-specific representations before enforcing strict cross-modal consistency.

Each branch is supervised by a task loss that enforces accurate position reconstruction and temporal smoothness. Let $\varphi \in \mathbb{R}^{N \times 6}$ denote the ground-truth trajectory, and $\hat{\varphi}_{b}$ denote the predicted trajectory from branch $b \in \{\sigma, \tau\}$. The task loss is defined as:
\begin{equation}
\mathcal{L}_{\text{task}}^{b} = \mathcal{L}_{\text{rec}}^b + \lambda_{\text{vel}} \mathcal{L}_{\text{vel}}^b,
\end{equation}
where the reconstruction loss $\mathcal{L}_{\text{rec}}^b$ and velocity loss $\mathcal{L}_{\text{vel}}^b$ are formulated as:
\begin{equation}
\mathcal{L}_{\text{rec}}^b = \frac{1}{N} \sum_{n=1}^{N} \|\hat{\varphi}_{b,n} - \varphi_n\|_2^2, 
\end{equation}
\begin{equation}
\mathcal{L}_{\text{vel}}^b = \frac{1}{N-1} \sum_{n=1}^{N-1} \|(\hat{\varphi}_{b, n+1} - \hat{\varphi}_{b, n}) - (\varphi_{n+1} - \varphi_{n})\|_2^2.
\end{equation}
Here, $\|\cdot\|_2^2$ denotes the squared $L_2$ norm, and $\lambda_{\text{vel}}$ is a hyperparameter controlling temporal smoothness.

The overall objective is:
\begin{equation}
\mathcal{L}_{\text{total}} = 
\mathcal{L}_{\text{task}}^{\sigma} 
+ \lambda_{\tau} \mathcal{L}_{\text{task}}^{\tau} 
+ \lambda_{\text{distill}} \mathcal{L}_{\text{align}},
\end{equation}
where $\lambda_{\tau}$ and $\lambda_{\text{distill}}$ control the contributions of the teacher supervision and distillation alignment, respectively. Ultimately, this stage effectively distills structured musical priors into the audio-driven student model while maintaining an efficient audio-only inference pipeline.

\subsection{Bimanual Coordinated Gesture Generation}
\label{sec:stage2}

This stage synthesizes fine-grained bimanual hand gestures $\mathbf{x}=[\mathbf{x}^L, \mathbf{x}^R] \in \mathbb{R}^{N \times D}$. As illustrated in Figure~\ref{fig:arch}, we adopt a flow-matching framework~\cite{lipman2022flow}, where a dual-stream spatio-temporal architecture independently predicts the continuous-time velocity fields for each hand. To bridge these unshared streams, a parameter-shared Asymmetric Role-Gated Interaction (ARGI) module explicitly models dynamic cross-hand coordination at the network bottleneck. During inference, we sample gestures by solving the Ordinary Differential Equation (ODE) induced by the learned velocity field.

\subsubsection{Velocity Field Network.}
\label{sec:stage2_arch}

The Velocity Field Network serves as the core generative backbone, designed to map the intermediate gesture state and conditioning features to a continuous-time velocity field.

\noindent\textbf{Conditioning.}
We construct a global conditioning feature $\mathbf{c} = [\mathbf{F}_{\text{audio}}, \mathbf{F}_{\sigma}^{\text{enc}}, \hat{\varphi}_{\sigma}]$, where $\mathbf{F}_{\text{audio}}$ denotes the frozen MuQ acoustic representation, $\mathbf{F}_{\sigma}^{\text{enc}}$ represents the student-side encoded feature, and $\hat{\varphi}_{\sigma}$ is the predicted bimanual wrist trajectory from Stage 1. At time $t \in [0, 1]$, the intermediate latent gesture state is denoted by $\mathbf{x}_t = [\mathbf{x}_t^L, \mathbf{x}_t^R]$.

\noindent\textbf{Factorized Spatio-Temporal Backbone.}
Specifically, a Spatial Transformer first models the intra-hand structural correlations by tokenizing individual kinematic joints to perform exhaustive intra-hand self-attention. Subsequently, a temporal 1D U-Net~\cite{ronneberger2015u} processes these aggregated spatial embeddings to capture multi-scale gesture dynamics along the temporal axis. The global conditioning $\mathbf{c}$ is injected into the intermediate features of each U-Net block via Feature-wise Linear Modulation~\cite{perez2018film}. At the network bottleneck, we extract the compressed representations $\mathbf{h}_s \in \mathbb{R}^{N' \times C}$ for each hand $s \in \{L, R\}$, serving as the foundation for the subsequent cross-hand interaction. Ultimately, the network outputs the conditional velocity field $\mathbf{v}_\theta(t, \mathbf{x}_t; \mathbf{c})$ for the concatenated bimanual representation.

\noindent\textbf{Asymmetric Role-Gated Interaction (ARGI).}
To explicitly model coordinated bimanual dynamics, we integrate cross-hand dependencies at the U-Net bottleneck. Given the bottleneck features $\mathbf{h}_s$, let $\bar{s}$ represent the contralateral hand. Unlike the branch-specific backbones, ARGI serves as a parameter-shared bridge to facilitate information exchange. To instill hand-specific awareness, the query $\mathbf{Q}_s^{\mathcal{I}}$ is derived by injecting a learnable hand identity embedding $\mathbf{e}_{\text{id}}(s)$ followed by a self-attention layer $\zeta_{\text{self}}$. Crucially, we introduce learnable projection matrices $\mathbf{W}_Q^{\mathcal{I}}, \mathbf{W}_K^{\mathcal{I}}, \mathbf{W}_V^{\mathcal{I}} \in \mathbb{R}^{C \times d_k}$ to map the representations into a dedicated interaction space:
\begin{equation}
\mathbf{Q}_s^{\mathcal{I}} = \zeta_{\text{self}}(\mathbf{h}_s + \mathbf{e}_{\text{id}}(s)) \mathbf{W}_Q^{\mathcal{I}}, \quad \mathbf{K}_{\bar{s}}^{\mathcal{I}} = \mathbf{h}_{\bar{s}} \mathbf{W}_K^{\mathcal{I}}, \quad \mathbf{V}_{\bar{s}}^{\mathcal{I}} = \mathbf{h}_{\bar{s}} \mathbf{W}_V^{\mathcal{I}}.
\label{eq:argi_qkv}
\end{equation}
The cross-hand message $\mathbf{m}_s$ is then synthesized via a cross-attention mechanism:
\begin{equation}
\mathbf{m}_s = \mathrm{Softmax}\left(\frac{\mathbf{Q}_s^{\mathcal{I}} (\mathbf{K}_{\bar{s}}^{\mathcal{I}})^\top}{\sqrt{d_k}}\right) \mathbf{V}_{\bar{s}}^{\mathcal{I}},
\label{eq:argi_attn}
\end{equation}
where $d_k$ denotes the channel dimension of the key space. To accommodate the dynamic nature of piano performance (e.g., shifts between melody and accompaniment roles), we employ a temporal gating mechanism to modulate the interaction strength. Conditioned on the local context of $\mathbf{h}_s$, a multi-layer perceptron (MLP) predicts a frame-wise gate $\mathbf{g}_s \in [0, 1]^{N'}$. The cross-hand message is then gated, refined by a feed-forward network (FFN), and integrated into the main flow:
\begin{equation}
\mathbf{g}_s = \sigma\big(\mathrm{MLP}(\mathbf{h}_s)\big), \quad \mathbf{h}^{\text{inter}}_s = \mathbf{h}_s + \mathrm{FFN}\big(\mathbf{g}_s \odot \mathbf{m}_s\big),
\label{eq:argi_gate}
\end{equation}
where $\sigma(\cdot)$ is the sigmoid function and $\odot$ denotes element-wise multiplication. The updated features $\mathbf{h}^{\text{inter}}_s$ are subsequently processed by the U-Net decoder to estimate the final velocity field. This design enables adaptive bimanual coordination by dynamically evaluating the implicit musical role of each hand at each time step.

\subsubsection{Flow Matching and Training Objective.}
\label{sec:stage2_fm}

To optimize the Velocity Field Network, we employ a conditional flow matching (CFM) framework, which provides a simulation-free regression objective for learning the continuous-time generative process.

\noindent\textbf{Conditional Flow Formulation.}
We learn a continuous-time conditional velocity field $\mathbf{v}_\theta(t, \mathbf{x}_t; \mathbf{c})$ that transports samples from a base distribution to the target data distribution, conditioned on the multi-modal features $\mathbf{c}$. During training, we draw a ground-truth gesture sample $\mathbf{x}_1 \sim p_{\text{data}}$, a noise sample $\mathbf{x}_0 \sim p_{\text{base}}$ from a standard Gaussian $\mathcal{N}(\mathbf{0}, \mathbf{I})$, and a time step $t \sim \mathcal{U}(0,1)$.

Following Optimal Transport (OT) principles~\cite{lipman2022flow,liu2022flow}, we define a linear probability path:
\begin{equation}
\mathbf{x}_t = (1-t)\mathbf{x}_0 + t\mathbf{x}_1,
\label{eq:prob_path}
\end{equation}
which induces a constant target velocity $\mathbf{u}_t = \mathbf{x}_1 - \mathbf{x}_0$. The network is trained to estimate the velocity field $\mathbf{v}_\theta(t, \mathbf{x}_t; \mathbf{c})$ by minimizing the conditional flow-matching objective:
\begin{equation}
\mathcal{L}_{\mathrm{FM}} = \mathbb{E}_{\mathbf{x}_0, \mathbf{x}_1, t} \left[ \left\| \mathbf{v}_\theta(t, \mathbf{x}_t; \mathbf{c}) - \mathbf{u}_t \right\|_2^2 \right].
\label{eq:fm_loss}
\end{equation}
During training, the ARGI module facilitates explicit cross-hand coupling within the network's intermediate layers, ensuring the synthesis of tightly coordinated gesture flows.

At inference time, gestures are generated by sampling an initial noise $\mathbf{x}_0 \sim p_{\text{base}}$ and solving the following Ordinary Differential Equation (ODE) from $t=0$ to $t=1$ using numerical solvers~\cite{chen2018neural}:
\begin{equation}
\frac{\mathrm{d}\mathbf{x}(t)}{\mathrm{d}t} = \mathbf{v}_\theta\big(t, \mathbf{x}(t); \mathbf{c}\big), \qquad \mathbf{x}(0) = \mathbf{x}_0.
\label{eq:ode_sampling}
\end{equation}
While standard adaptive-step ODE solvers are effective for fixed-length sequences, this formulation lacks the temporal continuity required for streaming synthesis of arbitrary length.

\subsection{Streaming Inference via Autoregressive Flow Continuation}
\label{sec:streaming_inference}

For arbitrary-length motion generation, we propose Autoregressive Flow Continuation (AFC). This method uses a causal sliding-window strategy to process gesture chunks sequentially. In this framework, the prior sequence provides temporal grounding for the current chunk via guided continuous-time inpainting in an overlap. 
To ensure a deterministic transport path, we fix a noise sample $\boldsymbol{\epsilon} \sim \mathcal{N}(\mathbf{0}, \mathbf{I})$. We denote the gesture from the preceding chunk's tail as $\mathbf{x}_{\text{past}}$, anchoring temporal continuity. During forward-time integration from $t$ to $t+\Delta t$, we obtain a provisional update $\mathbf{x}_{t+\Delta t}^{\text{uncond}}$ via a numerical ODE solver step:
\begin{equation}
\mathbf{x}_{t+\Delta t}^{\text{uncond}} = \Psi(\mathbf{x}_t, \mathbf{v}_\theta, t, \Delta t, \mathbf{c}),
\label{eq:afc_uncond}
\end{equation}
where $\Psi$ denotes the solver's transition function. Simultaneously, we construct a reference state $\mathbf{x}_{t+\Delta t}^{\text{known}}$ for the overlapping region by interpolating the fixed noise $\boldsymbol{\epsilon}$ with the known history $\mathbf{x}_{\text{past}}$ along the optimal transport path:
\begin{equation}
\mathbf{x}_{t+\Delta t}^{\text{known}} = (1 - (t + \Delta t))\boldsymbol{\epsilon} + (t + \Delta t)\mathbf{x}_{\text{past}}.
\label{eq:afc_known}
\end{equation}

We then refine the state by blending the known component with the provisional estimate using a temporal cosine mask $\mathbf{m}$ with values in $[0, 1]$:
\begin{equation}
\mathbf{x}_{t+\Delta t} \leftarrow \mathbf{m} \odot \mathbf{x}_{t+\Delta t}^{\text{known}} + (\mathbf{1} - \mathbf{m}) \odot \mathbf{x}_{t+\Delta t}^{\text{uncond}},
\label{eq:afc_blend}
\end{equation}
where $\mathbf{m}$ decays from $1$ to $0$ across the overlap region. This mechanism enforces boundary consistency while smoothing the transition into free generation. Executing this update at every solver iteration ensures that each new chunk integrates seamlessly with the prior gesture trajectory, maintaining long-term temporal coherence without inducing additional latency.

\begin{table*}[htbp]
\caption{Quantitative evaluation on the PianoMotion10M dataset under the standard short-sequence setting.}
\label{tab:quan}
\renewcommand\arraystretch{1.1}
\resizebox{1\linewidth}{!}{
\begin{tabular}{llcccccccccc}
\toprule
\multirow{2}{*}{\textbf{Method}} & \multirow{2}{*}{\textbf{Venue}} & \multicolumn{4}{c}{\textbf{Left Hand}}& \multicolumn{4}{c}{\textbf{Right Hand}}& \multirow{2}{*}{\textbf{FID$\downarrow$}} & \multirow{2}{*}{\textbf{RTF$\downarrow$}}\\ 
\cmidrule(lr){3-6} \cmidrule(lr){7-10}
 &  & \textbf{FGD$\downarrow$} & \textbf{WGD$\downarrow$} & \textbf{PD$\downarrow$} & \textbf{Smooth$\downarrow$} & \textbf{FGD$\downarrow$} & \textbf{WGD$\downarrow$} & \textbf{PD$\downarrow$} & \textbf{Smooth$\downarrow$} & & \\ 
\midrule
EmoTalk~\cite{peng2023emotalk} & ICCV 2023 & 0.445 & 0.232 & 0.044 & 0.353 & 0.360 & 0.259 & 0.033 & 0.313 & 4.645 & - \\
LivelySpeaker~\cite{zhi2023livelyspeaker} & ICCV 2023 & 0.538 & 0.220 & 0.038 & 0.406 & 0.535 & 0.249 & 0.030 & 0.334 & 4.157 & - \\ 
PianoMotion~\cite{pm} & ICLR 2025 & 0.372 & 0.217 & 0.037 & 0.248  & 0.351 & 0.244 & 0.030 & 0.205 & 3.281 & 1.746 \\ 
S2C~\cite{s2c} & MM 2025 & 0.309 & 0.215 & 0.035 & 0.229 & 0.293 & 0.242 & 0.028 & 0.198 & 3.012 & 5.723 \\ 
\midrule
PianoFlow (Ours) & - & \textbf{0.266} & \textbf{0.210} & \textbf{0.029} & \textbf{0.200} & \textbf{0.263} & \textbf{0.238} & \textbf{0.027} & \textbf{0.184} & \textbf{2.674} & \textbf{0.186} \\ 
\bottomrule
\end{tabular}
}
\end{table*}

\begin{table*}[htbp]
\centering 
\caption{Quantitative comparison of piano motion generation under the long-sequence setting on the PianoMotion10M dataset.}
\label{tab:quan_long}
\renewcommand\arraystretch{1.1} 
\begin{tabular}{lcccccccccc}
\toprule
\multirow{2}{*}{\textbf{Method}} & \multicolumn{5}{c}{\textbf{Left Hand}} & \multicolumn{5}{c}{\textbf{Right Hand}} \\ 
\cmidrule(lr){2-6} \cmidrule(lr){7-11}
 & \textbf{FGD$\downarrow$} & \textbf{WGD$\downarrow$} & \textbf{PD$\downarrow$} & \textbf{Smooth$\downarrow$} & \textbf{FDE$\downarrow$} & \textbf{FGD$\downarrow$} & \textbf{WGD$\downarrow$} & \textbf{PD$\downarrow$} & \textbf{Smooth$\downarrow$} & \textbf{FDE$\downarrow$} \\ 
\midrule

PianoMotion~\cite{pm} + SWF &0.432&\textbf{0.220}&0.045&0.253&0.334&0.424&0.250&0.037&0.208&0.269\\ 
S2C~\cite{s2c} + SWF &0.443&0.223&0.041&0.251&0.350&0.397&0.252&0.035&0.206&0.320\\ 
PianoFlow + SWF &0.317&0.223&\textbf{0.040}&0.236&0.324&0.296&0.251&\textbf{0.032}&0.203&0.272\\ 
PianoFlow + AFC&\textbf{0.309}&0.222&\textbf{0.040}&\textbf{0.232}&\textbf{0.312}&\textbf{0.290}&\textbf{0.249}&\textbf{0.032}&\textbf{0.199}&\textbf{0.255}\\ 
\bottomrule
\end{tabular}
\end{table*}
\section{Experiments}

\subsection{Experiments Setting}
\subsubsection{Datasets.}
To validate the effectiveness of PianoFlow, we conduct experiments on the PianoMotion10M dataset~\cite{pm}, comprising 1,966 piano performance sequences at 30~FPS, totaling 116 hours. The dataset features synchronized audio, MIDI, and hand motion annotations, comprising 3D wrist coordinates and hand poses parameterized by the MANO~\cite{mano} model. The raw data is partitioned into 30-second clips, yielding 7,519 training, 821 validation, and 8,399 test samples. In line with the configuration of Gan~\textit{et~al.}~\cite{pm}, we further extract 8-second segments for training and short-sequence evaluation.

\subsubsection{Evaluation Metrics.}
We evaluate the synthesized motions across three dimensions, including distributional fidelity, kinematic accuracy, and computational efficiency. This evaluation incorporates five base metrics~\cite{pm} along with two extended metrics.
(1) \textbf{Fréchet Inception Distance (FID)} quantifies the global distributional discrepancy between synthesized motions and ground-truth data within a pre-trained latent space. 
(2) \textbf{Fréchet Gesture Distance (FGD)} assesses hand-specific pose distributions by calculating the Fréchet distance within the MANO \cite{mano} parameter or joint coordinate space. 
(3) \textbf{Wasserstein Gesture Distance (WGD)} evaluates distributional alignment via optimal transport, calculating the Earth Mover's Distance (EMD) following PCA and Gaussian Mixture Model (GMM) fitting. 
(4) \textbf{Position Distance (PD)} quantifies wrist spatial accuracy via the mean Euclidean distance between predicted and ground-truth wrist coordinates across the sequence. 
(5) \textbf{Smoothness} evaluates physical plausibility by measuring acceleration discrepancies relative to the ground truth to ensure temporal stability. 
(6) \textbf{Real-Time Factor (RTF)} measures computational efficiency, where an $\text{RTF} < 1.0$ indicates real-time generation capability. 
(7) \textbf{Frequency Domain Error (FDE)} evaluates preservation of high-frequency kinematic details and penalizes over-smoothing artifacts from interpolation-based fusion. We compute amplitude spectra $S(f)$ and $\hat{S}(f)$ for the sequences via Fast Fourier Transform. Defining $\mathcal{H}$ as high-frequency bins exceeding $5$ Hz, FDE is the Mean Absolute Error (MAE) between spectral components to ensure precise motion alignment:
\begin{equation}
\text{FDE} = \frac{1}{|\mathcal{H}|} \sum_{f \in \mathcal{H}} \left| S(f) - \hat{S}(f) \right|.
\label{eq:fde}
\end{equation}
A lower FDE signifies superior retention of rapid musical articulations and spectral fidelity, ensuring synthesized motions are both energetic and rhythmically accurate.

For \textbf{long-sequence evaluation}, we implement a sliding-window protocol (window size of 240 frames, stride of 120) for FGD, WGD, PD, and Smoothness, reporting the averaged results across all windows.

\subsubsection{Baselines.}
We benchmark PianoFlow against several state-of-the-art baselines, encompassing both general-purpose motion synthesis frameworks and specialized piano-specific models. 
(1) \textbf{EmoTalk}~\cite{peng2023emotalk} employs a cross-reconstruction mechanism to learn disentangled latent spaces and utilizes a Transformer-based decoder for sequence generation. 
(2) \textbf{LivelySpeaker}~\cite{zhi2023livelyspeaker} is a two-stage framework that leverages pre-trained embeddings for conditioning and a diffusion-based refiner to synthesize expressive motions. 
(3) \textbf{PianoMotion}~\cite{pm} adopts a DDPM-based diffusion framework to jointly model bimanual motions. It generates initial trajectories as coarse guidance to condition a subsequent diffusion-based module for fingering synthesis.
(4) \textbf{S2C}~\cite{s2c} builds upon the framework of PianoMotion by adopting a dual-stream diffusion architecture to model bimanual motions independently. It introduces static interaction between the hands during the generation process.
Following Gan \textit{et al.}~\cite{pm}, EmoTalk and LivelySpeaker were redeployed and evaluated on the PianoMotion10M dataset.

To facilitate a consistent comparison in long-sequence synthesis, we extend these fixed-length baselines via a Sliding-Window Fusion (SWF) strategy. In this setup, we generate 8-second segments with a 1-second temporal overlap and apply linear cross-fading in the overlapping regions to ensure continuity across chunks.

\subsubsection{Implementation Details.}

In the first stage, the decoupled cross-modal distillation scheme is trained for 100K iterations with a batch size of 24, using loss weights $\lambda_{\text{vel}}=3.0$, $\lambda_{\tau}=1.0$, and $\lambda_{\text{distill}}=0.1$ to transfer harmonic knowledge from the multimodal teacher to the audio-only student.
In the second stage, the conditional flow-matching framework is trained for 250K iterations with a batch size of 16.
For both stages, we use 8-second sequence segments and the AdamW optimizer with an initial learning rate of $5 \times 10^{-5}$. A ReduceLROnPlateau scheduler (factor 0.5) is applied, with patience values of 10 and 15 for the first and second stages, respectively.
During inference, motion is generated using a second-order Heun solver~\cite{karras2022elucidating} with 25 sampling steps. All experiments are conducted on a single NVIDIA RTX 4090 GPU.

\begin{figure*}
    \centering
    \includegraphics[width=0.83\textwidth]{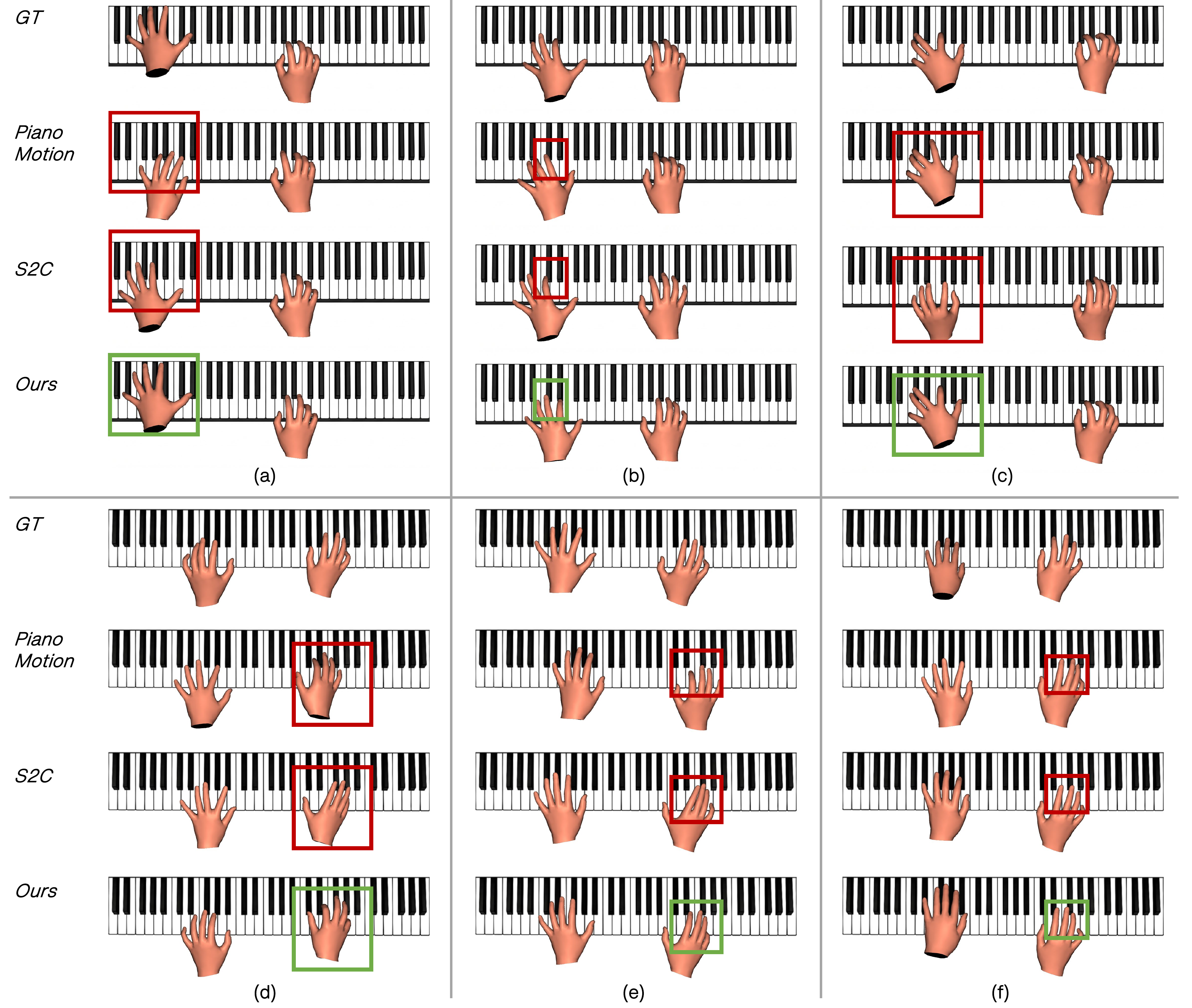}
    \caption{Qualitative comparison of synthesized motions, with red and green boxes highlighting kinematic inaccuracies and precise alignments, respectively. Overall, PianoFlow achieves superior spatial accuracy and bimanual coordination.}
    \label{fig:demo}
\end{figure*}

\subsection{Quantitative Results}

We quantitatively evaluate PianoFlow against established baselines for short- and long-sequence generation, with results detailed in Table~\ref{tab:quan} and Table~\ref{tab:quan_long}, respectively.
In the \textbf{short-sequence setting} presented in Table~\ref{tab:quan}, PianoFlow demonstrates superior performance across all evaluation metrics. Notably, it achieves significant improvements in global motion fidelity, reflected by the lowest FID of 2.674. When analyzing the left and right hands independently, our framework maintains high accuracy in both distributional similarity, indicated by the FGD and WGD metrics, and spatio-temporal dynamics, evidenced by the PD and Smoothness scores. Furthermore, PianoFlow exhibits exceptional computational efficiency with an RTF of 0.186, operating at least 9$\times$ faster than the baseline models. These results suggest that the synthesized motions closely resemble real performances while preserving spatial precision, temporal smoothness, and strict real-time execution capabilities.
For the \textbf{long-sequence generation} evaluated in Table~\ref{tab:quan_long}, PianoFlow maintains strong competitive performance, achieving the best results across almost all metrics. Notably, by achieving the best Smoothness and FDE scores, AFC successfully preserves rapid musical articulations and physical plausibility, effectively mitigating the over-smoothing artifacts inherent to interpolation-based SWF.

\subsection{Qualitative Results}

We qualitatively compare PianoFlow with competitive baselines, PianoMotion~\cite{pm} and S2C~\cite{s2c}. As illustrated in Figure~\ref{fig:demo}, PianoFlow generates more precise wrist trajectories and hand poses for accurate keystroke alignment. Specifically, Figure~\ref{fig:demo}(a) showcases a challenging wide-stretch gesture that our model synthesizes with higher fidelity compared to the baselines. 
Notably, Figure~\ref{fig:demo}(d) highlights PianoFlow's ability to capture the inherent flexibility of fingering strategies. Instead of rote memorization, the model explores diverse yet functionally equivalent fingering patterns to strike the same target keys. By synthesizing various biologically plausible poses in strict alignment with the audio, PianoFlow demonstrates its mastery of the keyboard's underlying physical constraints. Furthermore, our generated sequences maintain high temporal smoothness with minimal jitter throughout continuous performances, as demonstrated in the supplementary material.

\subsection{Ablation Study}

To evaluate the contribution of individual components, we categorize our ablation models into three dimensions:
(1) \textbf{Music Representations}: \textbf{w/ HuBERT} (replacing MuQ with HuBERT) and \textbf{w/o MIDI} (removing MIDI distillation).
(2) \textbf{Gesture Modeling}: \textbf{w/o Decoupling} (joint bimanual modeling in Stage 1), \textbf{w/o Stage 1} (removing wrist trajectory guidance), and \textbf{w/o Spatial} (omitting spatial Transformers).
(3) \textbf{Interaction Mechanisms}: \textbf{w/o ARGI} (independent hand modeling) and \textbf{w/ Cross-Attn} (replacing ARGI with vanilla cross-attention).
As shown in Table~\ref{tab:abla}, PianoFlow achieves the optimal balance across all metrics. The FID degradation in \textbf{w/ HuBERT} and \textbf{w/o MIDI} confirms that music-specific encoders and the use of a privileged modality are vital for capturing musical semantics. For gesture modeling, \textbf{w/o Stage 1} causes the left-hand PD to surge from 0.029 to 0.095, demonstrating that wrist trajectories serve as essential kinematic anchors. Although \textbf{w/o Spatial} achieves slightly better distributional alignment in terms of WGD, its Smoothness and FID noticeably degrade. These results and the performance drop in \textbf{w/o Decoupling} collectively justify our architecture design for preserving hand priors and joint correlations. Finally, ARGI outperforms independent modeling (\textbf{w/o ARGI}) and standard cross-attention (\textbf{w/ Cross-Attn}), validating that dynamic temporal gating effectively facilitates bimanual coordination while mitigating interference.

\begin{table*}[htbp]
\centering 
\caption{Ablation Study on the PianoMotion10M dataset under the standard short-sequence setting.}
\label{tab:abla}
\renewcommand\arraystretch{1.1} 
\begin{tabular}{lccccccccc}
\toprule
\multirow{2}{*}{\textbf{Method}} & \multicolumn{4}{c}{\textbf{Left Hand}} & \multicolumn{4}{c}{\textbf{Right Hand}} & \multirow{2}{*}{\textbf{FID$\downarrow$}} \\ 
\cmidrule(lr){2-5} \cmidrule(lr){6-9}
 & \textbf{FGD$\downarrow$} & \textbf{WGD$\downarrow$} & \textbf{PD$\downarrow$} & \textbf{Smooth$\downarrow$} & \textbf{FGD$\downarrow$} & \textbf{WGD$\downarrow$} & \textbf{PD$\downarrow$} & \textbf{Smooth$\downarrow$} & \\ 
\midrule

PianoFlow& \textbf{0.266} &  0.210 &  \textbf{0.029} &  \textbf{0.200} &  \textbf{0.263} & 0.238 & \textbf{0.027} & \textbf{0.184} & \textbf{2.674}\\ 
\quad w/ Hubert &0.293&0.214&0.032&0.251&0.286&0.240&0.029&0.196&2.868\\ 
\quad w/o MIDI &  0.277&0.211&0.030&0.203&0.281&0.241&0.028&0.194&2.733\\ 
\quad w/o Decoupling &0.283&0.211&0.038&0.221&0.293&0.241&0.035&0.198&2.877\\ 
\quad w/o Stage1 &  0.285 & 0.216 & 0.095 & 0.243 & 0.275 & 0.244 & 0.085 & 0.217 & 2.841 \\ 
\quad w/o Spatial &0.270& \textbf{0.209}&\textbf{0.029}&0.212&0.269& \textbf{0.237}& \textbf{0.027}&0.195&2.701\\ 
\quad w/o ARGI &0.280&0.212&\textbf{0.029}&0.228&0.288&0.238&\textbf{0.027}&0.196&2.796\\ 
\quad w/ Cross-Attn &0.283&0.211&\textbf{0.029}&0.222&0.286&0.240&\textbf{0.027}&0.191&2.778\\ 
\bottomrule
\end{tabular}
\end{table*}

\section{Applications}
\label{sec:applications}

PianoFlow demonstrates significant potential across various multimodal scenarios within the broader multimedia ecosystem. 

\noindent\textbf{AI-Assisted Music Education.} PianoFlow enhances pedagogical utility by converting audio into synchronized 3D demonstrations~\cite{xu2025study,pm}. By providing real-time gestural guidance from performance recordings, our framework offers scalable visual aids for mastering intricate fingerings and bimanual coordination, supplementing traditional instructor-led teaching with high-fidelity visual references.

\noindent\textbf{Interactive Systems and Virtual Livestreaming.} Leveraging the AFC mechanism, our framework supports low-latency interactive systems and virtual livestreaming~\cite{yang2025gesturehydra, chu2025artalk,yang2026streamingtalker}. It enables digital avatars to respond instantaneously to live musical inputs or audience prompts, ensuring precise synchronization for real-time virtual environments while addressing the latency bottlenecks prevalent in traditional offline-rendered animation.

\noindent\textbf{Controllable Performance Video Synthesis.} PianoFlow provides high-fidelity 3D gestural priors to bridge the modality gap in audio-to-video synthesis~\cite{gan2025omniavatar,chen2025x,lin2025omnihuman,wang2025universe}. By acting as a structural intermediary, these priors anchor downstream generative video models to physical reality, mitigating common artifacts such as unrealistic finger distortions and facilitating the synthesis of photorealistic, temporally coherent digital human performances.

\section{Conclusion}

We present \textbf{PianoFlow}, a novel two-stage framework for real-time, audio-driven bimanual piano motion generation. Through a cross-modal distillation scheme that leverages MIDI as privileged information during training, our model effectively captures deep harmonic structures while maintaining an efficient audio-only inference pipeline. To facilitate dynamic coordination, we develop the Asymmetric Role-Gated Interaction (ARGI) module to model time-varying bimanual dependencies. Furthermore, our Autoregressive Flow Continuation (AFC) mechanism enables seamless, temporally coherent streaming motion generation for sequences of arbitrary length. Extensive evaluations on the PianoMotion10M dataset demonstrate that PianoFlow establishes a new state-of-the-art in both motion fidelity and structural accuracy, while achieving inference speeds significantly faster than prior methods. Ultimately, PianoFlow offers a highly effective and scalable solution, paving the way for future advancements in digital human animation and real-time interactive virtual performances.

\bibliographystyle{ACM-Reference-Format}
\bibliography{sample-base}


\begin{thebibliography}{76}


\ifx \showCODEN    \undefined \def \showCODEN     #1{\unskip}     \fi
\ifx \showISBNx    \undefined \def \showISBNx     #1{\unskip}     \fi
\ifx \showISBNxiii \undefined \def \showISBNxiii  #1{\unskip}     \fi
\ifx \showISSN     \undefined \def \showISSN      #1{\unskip}     \fi
\ifx \showLCCN     \undefined \def \showLCCN      #1{\unskip}     \fi
\ifx \shownote     \undefined \def \shownote      #1{#1}          \fi
\ifx \showarticletitle \undefined \def \showarticletitle #1{#1}   \fi
\ifx \showURL      \undefined \def \showURL       {\relax}        \fi
\providecommand\bibfield[2]{#2}
\providecommand\bibinfo[2]{#2}
\providecommand\natexlab[1]{#1}
\providecommand\showeprint[2][]{arXiv:#2}

\bibitem[Aneja et~al\mbox{.}(2024)]%
        {aneja2024facetalk}
\bibfield{author}{\bibinfo{person}{Shivangi Aneja}, \bibinfo{person}{Justus Thies}, \bibinfo{person}{Angela Dai}, {and} \bibinfo{person}{Matthias Nie{\ss}ner}.} \bibinfo{year}{2024}\natexlab{}.
\newblock \showarticletitle{Facetalk: Audio-driven motion diffusion for neural parametric head models}. In \bibinfo{booktitle}{\emph{Proceedings of the IEEE/CVF conference on computer vision and pattern recognition}}. \bibinfo{pages}{21263--21273}.
\newblock


\bibitem[Aslam et~al\mbox{.}(2023)]%
        {aslam2023privileged}
\bibfield{author}{\bibinfo{person}{Muhammad~Haseeb Aslam}, \bibinfo{person}{Muhammad~Osama Zeeshan}, \bibinfo{person}{Marco Pedersoli}, \bibinfo{person}{Alessandro~L Koerich}, \bibinfo{person}{Simon Bacon}, {and} \bibinfo{person}{Eric Granger}.} \bibinfo{year}{2023}\natexlab{}.
\newblock \showarticletitle{Privileged knowledge distillation for dimensional emotion recognition in the wild}. In \bibinfo{booktitle}{\emph{Proceedings of the IEEE/CVF conference on computer vision and pattern recognition}}. \bibinfo{pages}{3338--3347}.
\newblock


\bibitem[Baevski et~al\mbox{.}(2020)]%
        {baevski2020wav2vec}
\bibfield{author}{\bibinfo{person}{Alexei Baevski}, \bibinfo{person}{Yuhao Zhou}, \bibinfo{person}{Abdelrahman Mohamed}, {and} \bibinfo{person}{Michael Auli}.} \bibinfo{year}{2020}\natexlab{}.
\newblock \showarticletitle{wav2vec 2.0: A framework for self-supervised learning of speech representations}.
\newblock \bibinfo{journal}{\emph{Advances in neural information processing systems}}  \bibinfo{volume}{33} (\bibinfo{year}{2020}), \bibinfo{pages}{12449--12460}.
\newblock


\bibitem[Cassell et~al\mbox{.}(2001)]%
        {cassell2001beat}
\bibfield{author}{\bibinfo{person}{Justine Cassell}, \bibinfo{person}{Hannes~H{\"o}gni Vilhj{\'a}lmsson}, {and} \bibinfo{person}{Timothy Bickmore}.} \bibinfo{year}{2001}\natexlab{}.
\newblock \showarticletitle{Beat: the behavior expression animation toolkit}. In \bibinfo{booktitle}{\emph{Proceedings of the 28th annual conference on Computer graphics and interactive techniques}}. \bibinfo{pages}{477--486}.
\newblock


\bibitem[Chae-Yeon et~al\mbox{.}(2025)]%
        {chae2025perceptually}
\bibfield{author}{\bibinfo{person}{Lee Chae-Yeon}, \bibinfo{person}{Oh Hyun-Bin}, \bibinfo{person}{Han EunGi}, \bibinfo{person}{Kim Sung-Bin}, \bibinfo{person}{Suekyeong Nam}, {and} \bibinfo{person}{Tae-Hyun Oh}.} \bibinfo{year}{2025}\natexlab{}.
\newblock \showarticletitle{Perceptually accurate 3d talking head generation: New definitions, speech-mesh representation, and evaluation metrics}. In \bibinfo{booktitle}{\emph{Proceedings of the IEEE/CVF Conference on Computer Vision and Pattern Recognition}}. \bibinfo{pages}{21065--21074}.
\newblock


\bibitem[Chen et~al\mbox{.}(2025b)]%
        {chen2025language}
\bibfield{author}{\bibinfo{person}{Changan Chen}, \bibinfo{person}{Juze Zhang}, \bibinfo{person}{Shrinidhi~K Lakshmikanth}, \bibinfo{person}{Yusu Fang}, \bibinfo{person}{Ruizhi Shao}, \bibinfo{person}{Gordon Wetzstein}, \bibinfo{person}{Li Fei-Fei}, {and} \bibinfo{person}{Ehsan Adeli}.} \bibinfo{year}{2025}\natexlab{b}.
\newblock \showarticletitle{The language of motion: Unifying verbal and non-verbal language of 3d human motion}. In \bibinfo{booktitle}{\emph{Proceedings of the Computer Vision and Pattern Recognition Conference}}. \bibinfo{pages}{6200--6211}.
\newblock


\bibitem[Chen et~al\mbox{.}(2021)]%
        {chen2021music}
\bibfield{author}{\bibinfo{person}{Jiali Chen}, \bibinfo{person}{Changjie Fan}, \bibinfo{person}{Zhimeng Zhang}, \bibinfo{person}{Gongzheng Li}, \bibinfo{person}{Zeng Zhao}, \bibinfo{person}{Zhigang Deng}, {and} \bibinfo{person}{Yu Ding}.} \bibinfo{year}{2021}\natexlab{}.
\newblock \showarticletitle{A music-driven deep generative adversarial model for guzheng playing animation}.
\newblock \bibinfo{journal}{\emph{IEEE Transactions on Visualization and Computer Graphics}} \bibinfo{volume}{29}, \bibinfo{number}{2} (\bibinfo{year}{2021}), \bibinfo{pages}{1400--1414}.
\newblock


\bibitem[Chen et~al\mbox{.}(2018)]%
        {chen2018neural}
\bibfield{author}{\bibinfo{person}{Ricky~TQ Chen}, \bibinfo{person}{Yulia Rubanova}, \bibinfo{person}{Jesse Bettencourt}, {and} \bibinfo{person}{David~K Duvenaud}.} \bibinfo{year}{2018}\natexlab{}.
\newblock \showarticletitle{Neural ordinary differential equations}.
\newblock \bibinfo{journal}{\emph{Advances in neural information processing systems}}  \bibinfo{volume}{31} (\bibinfo{year}{2018}).
\newblock


\bibitem[Chen et~al\mbox{.}(2025a)]%
        {chen2025x}
\bibfield{author}{\bibinfo{person}{Zeyuan Chen}, \bibinfo{person}{Hongyi Xu}, \bibinfo{person}{Guoxian Song}, \bibinfo{person}{You Xie}, \bibinfo{person}{Chenxu Zhang}, \bibinfo{person}{Xin Chen}, \bibinfo{person}{Chao Wang}, \bibinfo{person}{Di Chang}, {and} \bibinfo{person}{Linjie Luo}.} \bibinfo{year}{2025}\natexlab{a}.
\newblock \showarticletitle{X-dancer: Expressive music to human dance video generation}. In \bibinfo{booktitle}{\emph{Proceedings of the IEEE/CVF International Conference on Computer Vision}}. \bibinfo{pages}{10602--10611}.
\newblock


\bibitem[Chhatre et~al\mbox{.}(2024)]%
        {chhatre2024emotional}
\bibfield{author}{\bibinfo{person}{Kiran Chhatre}, \bibinfo{person}{Nikos Athanasiou}, \bibinfo{person}{Giorgio Becherini}, \bibinfo{person}{Christopher Peters}, \bibinfo{person}{Michael~J Black}, \bibinfo{person}{Timo Bolkart}, {et~al\mbox{.}}} \bibinfo{year}{2024}\natexlab{}.
\newblock \showarticletitle{Emotional speech-driven 3d body animation via disentangled latent diffusion}. In \bibinfo{booktitle}{\emph{Proceedings of the IEEE/CVF Conference on Computer Vision and Pattern Recognition}}. \bibinfo{pages}{1942--1953}.
\newblock


\bibitem[Chu et~al\mbox{.}(2025)]%
        {chu2025artalk}
\bibfield{author}{\bibinfo{person}{Xuangeng Chu}, \bibinfo{person}{Nabarun Goswami}, \bibinfo{person}{Ziteng Cui}, \bibinfo{person}{Hanqin Wang}, {and} \bibinfo{person}{Tatsuya Harada}.} \bibinfo{year}{2025}\natexlab{}.
\newblock \showarticletitle{Artalk: Speech-driven 3d head animation via autoregressive model}. In \bibinfo{booktitle}{\emph{Proceedings of the SIGGRAPH Asia 2025 Conference Papers}}. \bibinfo{pages}{1--9}.
\newblock


\bibitem[Dai et~al\mbox{.}(2026)]%
        {dai2026tcdif}
\bibfield{author}{\bibinfo{person}{Yuqin Dai}, \bibinfo{person}{Wanlu Zhu}, \bibinfo{person}{Ronghui Li}, \bibinfo{person}{Xiu Li}, \bibinfo{person}{Zhenyu Zhang}, \bibinfo{person}{Jun Li}, {and} \bibinfo{person}{Jian Yang}.} \bibinfo{year}{2026}\natexlab{}.
\newblock \showarticletitle{TCDiff++: An End-to-end Trajectory-Controllable Diffusion Model for Harmonious Music-Driven Group Choreography}.
\newblock \bibinfo{journal}{\emph{International Journal of Computer Vision}} \bibinfo{volume}{134}, \bibinfo{number}{2} (\bibinfo{year}{2026}), \bibinfo{pages}{61}.
\newblock


\bibitem[Fan et~al\mbox{.}(2025)]%
        {fan2025align}
\bibfield{author}{\bibinfo{person}{Congyi Fan}, \bibinfo{person}{Jian Guan}, \bibinfo{person}{Xuanjia Zhao}, \bibinfo{person}{Dongli Xu}, \bibinfo{person}{Youtian Lin}, \bibinfo{person}{Tong Ye}, \bibinfo{person}{Pengming Feng}, {and} \bibinfo{person}{Haiwei Pan}.} \bibinfo{year}{2025}\natexlab{}.
\newblock \showarticletitle{Align your rhythm: Generating highly aligned dance poses with gating-enhanced rhythm-aware feature representation}. In \bibinfo{booktitle}{\emph{Proceedings of the IEEE/CVF International Conference on Computer Vision}}. \bibinfo{pages}{13193--13202}.
\newblock


\bibitem[Fan et~al\mbox{.}(2024)]%
        {fan2024unitalker}
\bibfield{author}{\bibinfo{person}{Xiangyu Fan}, \bibinfo{person}{Jiaqi Li}, \bibinfo{person}{Zhiqian Lin}, \bibinfo{person}{Weiye Xiao}, {and} \bibinfo{person}{Lei Yang}.} \bibinfo{year}{2024}\natexlab{}.
\newblock \showarticletitle{Unitalker: Scaling up audio-driven 3d facial animation through a unified model}. In \bibinfo{booktitle}{\emph{European Conference on Computer Vision}}. Springer, \bibinfo{pages}{204--221}.
\newblock


\bibitem[Fan et~al\mbox{.}(2022)]%
        {fan2022faceformer}
\bibfield{author}{\bibinfo{person}{Yingruo Fan}, \bibinfo{person}{Zhaojiang Lin}, \bibinfo{person}{Jun Saito}, \bibinfo{person}{Wenping Wang}, {and} \bibinfo{person}{Taku Komura}.} \bibinfo{year}{2022}\natexlab{}.
\newblock \showarticletitle{Faceformer: Speech-driven 3d facial animation with transformers}. In \bibinfo{booktitle}{\emph{Proceedings of the IEEE/CVF conference on computer vision and pattern recognition}}. \bibinfo{pages}{18770--18780}.
\newblock


\bibitem[Gan et~al\mbox{.}(2024)]%
        {pm}
\bibfield{author}{\bibinfo{person}{Qijun Gan}, \bibinfo{person}{Song Wang}, \bibinfo{person}{Shengtao Wu}, {and} \bibinfo{person}{Jianke Zhu}.} \bibinfo{year}{2024}\natexlab{}.
\newblock \showarticletitle{PianoMotion10M: Dataset and Benchmark for Hand Motion Generation in Piano Performance}.
\newblock \bibinfo{journal}{\emph{arXiv preprint arXiv:2406.09326}} (\bibinfo{year}{2024}).
\newblock


\bibitem[Gan et~al\mbox{.}(2025)]%
        {gan2025omniavatar}
\bibfield{author}{\bibinfo{person}{Qijun Gan}, \bibinfo{person}{Ruizi Yang}, \bibinfo{person}{Jianke Zhu}, \bibinfo{person}{Shaofei Xue}, {and} \bibinfo{person}{Steven Hoi}.} \bibinfo{year}{2025}\natexlab{}.
\newblock \showarticletitle{Omniavatar: Efficient audio-driven avatar video generation with adaptive body animation}.
\newblock \bibinfo{journal}{\emph{arXiv preprint arXiv:2506.18866}} (\bibinfo{year}{2025}).
\newblock


\bibitem[Ghosh et~al\mbox{.}(2025)]%
        {ghosh2025duetgen}
\bibfield{author}{\bibinfo{person}{Anindita Ghosh}, \bibinfo{person}{Bing Zhou}, \bibinfo{person}{Rishabh Dabral}, \bibinfo{person}{Jian Wang}, \bibinfo{person}{Vladislav Golyanik}, \bibinfo{person}{Christian Theobalt}, \bibinfo{person}{Philipp Slusallek}, {and} \bibinfo{person}{Chuan Guo}.} \bibinfo{year}{2025}\natexlab{}.
\newblock \showarticletitle{Duetgen: Music driven two-person dance generation via hierarchical masked modeling}. In \bibinfo{booktitle}{\emph{Proceedings of the Special Interest Group on Computer Graphics and Interactive Techniques Conference Conference Papers}}. \bibinfo{pages}{1--11}.
\newblock


\bibitem[Guo et~al\mbox{.}(2025)]%
        {guo2025controllable}
\bibfield{author}{\bibinfo{person}{Puyuan Guo}, \bibinfo{person}{Tuo Hao}, \bibinfo{person}{Wenxin Fu}, \bibinfo{person}{Yingming Gao}, {and} \bibinfo{person}{Ya Li}.} \bibinfo{year}{2025}\natexlab{}.
\newblock \showarticletitle{Controllable 3d dance generation using diffusion-based transformer u-net}. In \bibinfo{booktitle}{\emph{Proceedings of the AAAI Conference on Artificial Intelligence}}, Vol.~\bibinfo{volume}{39}. \bibinfo{pages}{3284--3292}.
\newblock


\bibitem[Hsu et~al\mbox{.}(2021)]%
        {hsu2021hubert}
\bibfield{author}{\bibinfo{person}{Wei-Ning Hsu}, \bibinfo{person}{Benjamin Bolte}, \bibinfo{person}{Yao-Hung~Hubert Tsai}, \bibinfo{person}{Kushal Lakhotia}, \bibinfo{person}{Ruslan Salakhutdinov}, {and} \bibinfo{person}{Abdelrahman Mohamed}.} \bibinfo{year}{2021}\natexlab{}.
\newblock \showarticletitle{Hu{BERT}: Self-supervised speech representation learning by masked prediction of hidden units}.
\newblock \bibinfo{journal}{\emph{IEEE/ACM transactions on audio, speech, and language processing}}  \bibinfo{volume}{29} (\bibinfo{year}{2021}), \bibinfo{pages}{3451--3460}.
\newblock


\bibitem[Huang et~al\mbox{.}(2024)]%
        {huang2024beat}
\bibfield{author}{\bibinfo{person}{Zikai Huang}, \bibinfo{person}{Xuemiao Xu}, \bibinfo{person}{Cheng Xu}, \bibinfo{person}{Huaidong Zhang}, \bibinfo{person}{Chenxi Zheng}, \bibinfo{person}{Jing Qin}, {and} \bibinfo{person}{Shengfeng He}.} \bibinfo{year}{2024}\natexlab{}.
\newblock \showarticletitle{Beat-it: Beat-synchronized multi-condition 3d dance generation}. In \bibinfo{booktitle}{\emph{European conference on computer vision}}. Springer, \bibinfo{pages}{273--290}.
\newblock


\bibitem[Jaegle et~al\mbox{.}(2021)]%
        {Jaegle2021Perceiver}
\bibfield{author}{\bibinfo{person}{Andrew Jaegle}, \bibinfo{person}{Felix Gimeno}, \bibinfo{person}{Andy Brock}, \bibinfo{person}{Oriol Vinyals}, \bibinfo{person}{Andrew Zisserman}, {and} \bibinfo{person}{Joao Carreira}.} \bibinfo{year}{2021}\natexlab{}.
\newblock \showarticletitle{Perceiver: General perception with iterative attention}. In \bibinfo{booktitle}{\emph{International conference on machine learning}}. PMLR, \bibinfo{pages}{4651--4664}.
\newblock


\bibitem[Jafari et~al\mbox{.}(2024)]%
        {jafari2024jambatalk}
\bibfield{author}{\bibinfo{person}{Farzaneh Jafari}, \bibinfo{person}{Stefano Berretti}, {and} \bibinfo{person}{Anup Basu}.} \bibinfo{year}{2024}\natexlab{}.
\newblock \showarticletitle{JambaTalk: Speech-driven 3D Talking Head Generation based on a Hybrid Transformer-Mamba Model}.
\newblock \bibinfo{journal}{\emph{ACM Transactions on Multimedia Computing, Communications and Applications}} (\bibinfo{year}{2024}).
\newblock


\bibitem[Jiang et~al\mbox{.}(2024)]%
        {jiang2024audio}
\bibfield{author}{\bibinfo{person}{Diqiong Jiang}, \bibinfo{person}{Jian Chang}, \bibinfo{person}{Lihua You}, \bibinfo{person}{Shaojun Bian}, \bibinfo{person}{Robert Kosk}, {and} \bibinfo{person}{Greg Maguire}.} \bibinfo{year}{2024}\natexlab{}.
\newblock \showarticletitle{Audio-driven facial animation with deep learning: A survey}.
\newblock \bibinfo{journal}{\emph{Information}} \bibinfo{volume}{15}, \bibinfo{number}{11} (\bibinfo{year}{2024}), \bibinfo{pages}{675}.
\newblock


\bibitem[Jiao et~al\mbox{.}(2025)]%
        {jiao2025bach}
\bibfield{author}{\bibinfo{person}{Jihui Jiao}, \bibinfo{person}{Rui Zeng}, \bibinfo{person}{Ju Dai}, {and} \bibinfo{person}{Junjun Pan}.} \bibinfo{year}{2025}\natexlab{}.
\newblock \showarticletitle{BACH: Bi-Stage Data-Driven Piano Performance Animation for Controllable Hand Motion}.
\newblock \bibinfo{journal}{\emph{Computer Animation and Virtual Worlds}} \bibinfo{volume}{36}, \bibinfo{number}{3} (\bibinfo{year}{2025}), \bibinfo{pages}{e70044}.
\newblock


\bibitem[Jin et~al\mbox{.}(2024)]%
        {jin2024audio}
\bibfield{author}{\bibinfo{person}{Yitong Jin}, \bibinfo{person}{Zhiping Qiu}, \bibinfo{person}{Yi Shi}, \bibinfo{person}{Shuangpeng Sun}, \bibinfo{person}{Chongwu Wang}, \bibinfo{person}{Donghao Pan}, \bibinfo{person}{Jiachen Zhao}, \bibinfo{person}{Zhenghao Liang}, \bibinfo{person}{Yuan Wang}, \bibinfo{person}{Xiaobing Li}, {et~al\mbox{.}}} \bibinfo{year}{2024}\natexlab{}.
\newblock \showarticletitle{Audio matters too! enhancing markerless motion capture with audio signals for string performance capture}.
\newblock \bibinfo{journal}{\emph{ACM Transactions on Graphics (TOG)}} \bibinfo{volume}{43}, \bibinfo{number}{4} (\bibinfo{year}{2024}), \bibinfo{pages}{1--10}.
\newblock


\bibitem[Karras et~al\mbox{.}(2022)]%
        {karras2022elucidating}
\bibfield{author}{\bibinfo{person}{Tero Karras}, \bibinfo{person}{Miika Aittala}, \bibinfo{person}{Timo Aila}, {and} \bibinfo{person}{Samuli Laine}.} \bibinfo{year}{2022}\natexlab{}.
\newblock \showarticletitle{Elucidating the design space of diffusion-based generative models}.
\newblock \bibinfo{journal}{\emph{Advances in neural information processing systems}}  \bibinfo{volume}{35} (\bibinfo{year}{2022}), \bibinfo{pages}{26565--26577}.
\newblock


\bibitem[Kim et~al\mbox{.}(2025b)]%
        {kim2025memorytalker}
\bibfield{author}{\bibinfo{person}{Hyung~Kyu Kim}, \bibinfo{person}{Sangmin Lee}, {and} \bibinfo{person}{Hak~Gu Kim}.} \bibinfo{year}{2025}\natexlab{b}.
\newblock \showarticletitle{MemoryTalker: Personalized Speech-Driven 3D Facial Animation via Audio-Guided Stylization}. In \bibinfo{booktitle}{\emph{Proceedings of the IEEE/CVF International Conference on Computer Vision}}. \bibinfo{pages}{11241--11251}.
\newblock


\bibitem[Kim et~al\mbox{.}(2025a)]%
        {kim2025deeptalk}
\bibfield{author}{\bibinfo{person}{Jisoo Kim}, \bibinfo{person}{Jungbin Cho}, \bibinfo{person}{Joonho Park}, \bibinfo{person}{Soonmin Hwang}, \bibinfo{person}{Da~Eun Kim}, \bibinfo{person}{Geon Kim}, {and} \bibinfo{person}{Youngjae Yu}.} \bibinfo{year}{2025}\natexlab{a}.
\newblock \showarticletitle{Deeptalk: Dynamic emotion embedding for probabilistic speech-driven 3d face animation}. In \bibinfo{booktitle}{\emph{Proceedings of the AAAI conference on artificial intelligence}}, Vol.~\bibinfo{volume}{39}. \bibinfo{pages}{4275--4283}.
\newblock


\bibitem[Kim et~al\mbox{.}(2022)]%
        {kim2022brand}
\bibfield{author}{\bibinfo{person}{Jinwoo Kim}, \bibinfo{person}{Heeseok Oh}, \bibinfo{person}{Seongjean Kim}, \bibinfo{person}{Hoseok Tong}, {and} \bibinfo{person}{Sanghoon Lee}.} \bibinfo{year}{2022}\natexlab{}.
\newblock \showarticletitle{A brand new dance partner: Music-conditioned pluralistic dancing controlled by multiple dance genres}. In \bibinfo{booktitle}{\emph{Proceedings of the IEEE/CVF Conference on Computer Vision and Pattern Recognition}}. \bibinfo{pages}{3490--3500}.
\newblock


\bibitem[Li et~al\mbox{.}(2024)]%
        {li2024lodge}
\bibfield{author}{\bibinfo{person}{Ronghui Li}, \bibinfo{person}{YuXiang Zhang}, \bibinfo{person}{Yachao Zhang}, \bibinfo{person}{Hongwen Zhang}, \bibinfo{person}{Jie Guo}, \bibinfo{person}{Yan Zhang}, \bibinfo{person}{Yebin Liu}, {and} \bibinfo{person}{Xiu Li}.} \bibinfo{year}{2024}\natexlab{}.
\newblock \showarticletitle{Lodge: A coarse to fine diffusion network for long dance generation guided by the characteristic dance primitives}. In \bibinfo{booktitle}{\emph{Proceedings of the IEEE/CVF Conference on Computer Vision and Pattern Recognition}}. \bibinfo{pages}{1524--1534}.
\newblock


\bibitem[Li et~al\mbox{.}(2023)]%
        {li2023finedance}
\bibfield{author}{\bibinfo{person}{Ronghui Li}, \bibinfo{person}{Junfan Zhao}, \bibinfo{person}{Yachao Zhang}, \bibinfo{person}{Mingyang Su}, \bibinfo{person}{Zeping Ren}, \bibinfo{person}{Han Zhang}, \bibinfo{person}{Yansong Tang}, {and} \bibinfo{person}{Xiu Li}.} \bibinfo{year}{2023}\natexlab{}.
\newblock \showarticletitle{Finedance: A fine-grained choreography dataset for 3d full body dance generation}. In \bibinfo{booktitle}{\emph{Proceedings of the IEEE/CVF International Conference on Computer Vision}}. \bibinfo{pages}{10234--10243}.
\newblock


\bibitem[Li et~al\mbox{.}(2025)]%
        {li2025music}
\bibfield{author}{\bibinfo{person}{Xiaojie Li}, \bibinfo{person}{Ronghui Li}, \bibinfo{person}{Shukai Fang}, \bibinfo{person}{Shuzhao Xie}, \bibinfo{person}{Xiaoyang Guo}, \bibinfo{person}{Jiaqing Zhou}, \bibinfo{person}{Junkun Peng}, {and} \bibinfo{person}{Zhi Wang}.} \bibinfo{year}{2025}\natexlab{}.
\newblock \showarticletitle{Music-aligned holistic 3d dance generation via hierarchical motion modeling}. In \bibinfo{booktitle}{\emph{Proceedings of the IEEE/CVF International Conference on Computer Vision}}. \bibinfo{pages}{14420--14430}.
\newblock


\bibitem[Lin et~al\mbox{.}(2025)]%
        {lin2025omnihuman}
\bibfield{author}{\bibinfo{person}{Gaojie Lin}, \bibinfo{person}{Jianwen Jiang}, \bibinfo{person}{Jiaqi Yang}, \bibinfo{person}{Zerong Zheng}, \bibinfo{person}{Chao Liang}, \bibinfo{person}{Yuan Zhang}, {and} \bibinfo{person}{Jingtuo Liu}.} \bibinfo{year}{2025}\natexlab{}.
\newblock \showarticletitle{Omnihuman-1: Rethinking the scaling-up of one-stage conditioned human animation models}. In \bibinfo{booktitle}{\emph{Proceedings of the IEEE/CVF International Conference on Computer Vision}}. \bibinfo{pages}{13847--13858}.
\newblock


\bibitem[Lipman et~al\mbox{.}(2022)]%
        {lipman2022flow}
\bibfield{author}{\bibinfo{person}{Yaron Lipman}, \bibinfo{person}{Ricky~TQ Chen}, \bibinfo{person}{Heli Ben-Hamu}, \bibinfo{person}{Maximilian Nickel}, {and} \bibinfo{person}{Matt Le}.} \bibinfo{year}{2022}\natexlab{}.
\newblock \showarticletitle{Flow matching for generative modeling}.
\newblock \bibinfo{journal}{\emph{arXiv preprint arXiv:2210.02747}} (\bibinfo{year}{2022}).
\newblock


\bibitem[Liu et~al\mbox{.}(2022b)]%
        {liu2022disco}
\bibfield{author}{\bibinfo{person}{Haiyang Liu}, \bibinfo{person}{Naoya Iwamoto}, \bibinfo{person}{Zihao Zhu}, \bibinfo{person}{Zhengqing Li}, \bibinfo{person}{You Zhou}, \bibinfo{person}{Elif Bozkurt}, {and} \bibinfo{person}{Bo Zheng}.} \bibinfo{year}{2022}\natexlab{b}.
\newblock \showarticletitle{Disco: Disentangled implicit content and rhythm learning for diverse co-speech gestures synthesis}. In \bibinfo{booktitle}{\emph{Proceedings of the 30th ACM international conference on multimedia}}. \bibinfo{pages}{3764--3773}.
\newblock


\bibitem[Liu et~al\mbox{.}(2024b)]%
        {liu2024emage}
\bibfield{author}{\bibinfo{person}{Haiyang Liu}, \bibinfo{person}{Zihao Zhu}, \bibinfo{person}{Giorgio Becherini}, \bibinfo{person}{Yichen Peng}, \bibinfo{person}{Mingyang Su}, \bibinfo{person}{You Zhou}, \bibinfo{person}{Xuefei Zhe}, \bibinfo{person}{Naoya Iwamoto}, \bibinfo{person}{Bo Zheng}, {and} \bibinfo{person}{Michael~J Black}.} \bibinfo{year}{2024}\natexlab{b}.
\newblock \showarticletitle{Emage: Towards unified holistic co-speech gesture generation via expressive masked audio gesture modeling}. In \bibinfo{booktitle}{\emph{Proceedings of the IEEE/CVF conference on computer vision and pattern recognition}}. \bibinfo{pages}{1144--1154}.
\newblock


\bibitem[Liu et~al\mbox{.}(2025b)]%
        {liu2025semges}
\bibfield{author}{\bibinfo{person}{Lanmiao Liu}, \bibinfo{person}{Esam Ghaleb}, \bibinfo{person}{Asli Ozyurek}, {and} \bibinfo{person}{Zerrin Yumak}.} \bibinfo{year}{2025}\natexlab{b}.
\newblock \showarticletitle{SemGes: Semantics-aware co-speech gesture generation using semantic coherence and relevance learning}. In \bibinfo{booktitle}{\emph{Proceedings of the IEEE/CVF International Conference on Computer Vision}}. \bibinfo{pages}{13963--13973}.
\newblock


\bibitem[Liu et~al\mbox{.}(2025a)]%
        {liu2025gcdance}
\bibfield{author}{\bibinfo{person}{Xinran Liu}, \bibinfo{person}{Xu Dong}, \bibinfo{person}{Shenbin Qian}, \bibinfo{person}{Diptesh Kanojia}, \bibinfo{person}{Wenwu Wang}, {and} \bibinfo{person}{Zhenhua Feng}.} \bibinfo{year}{2025}\natexlab{a}.
\newblock \showarticletitle{GCDance: Genre-Controlled Music-Driven 3D Full Body Dance Generation}.
\newblock \bibinfo{journal}{\emph{arXiv preprint arXiv:2502.18309}} (\bibinfo{year}{2025}).
\newblock


\bibitem[Liu et~al\mbox{.}({[n.\,d.]})]%
        {liu2024dgfm}
\bibfield{author}{\bibinfo{person}{Xinran Liu}, \bibinfo{person}{Zhenhua Feng}, \bibinfo{person}{Diptesh Kanojia}, {and} \bibinfo{person}{Wenwu Wang}.} \bibinfo{year}{[n.\,d.]}\natexlab{}.
\newblock \showarticletitle{DGFM: Full Body Dance Generation Driven by Music Foundation Models}. In \bibinfo{booktitle}{\emph{Audio Imagination: NeurIPS 2024 Workshop AI-Driven Speech, Music, and Sound Generation}}.
\newblock


\bibitem[Liu et~al\mbox{.}(2022a)]%
        {liu2022flow}
\bibfield{author}{\bibinfo{person}{Xingchao Liu}, \bibinfo{person}{Chengyue Gong}, {and} \bibinfo{person}{Qiang Liu}.} \bibinfo{year}{2022}\natexlab{a}.
\newblock \showarticletitle{Flow straight and fast: Learning to generate and transfer data with rectified flow}.
\newblock \bibinfo{journal}{\emph{arXiv preprint arXiv:2209.03003}} (\bibinfo{year}{2022}).
\newblock


\bibitem[Liu et~al\mbox{.}(2024a)]%
        {liu2024towards}
\bibfield{author}{\bibinfo{person}{Yifei Liu}, \bibinfo{person}{Qiong Cao}, \bibinfo{person}{Yandong Wen}, \bibinfo{person}{Huaiguang Jiang}, {and} \bibinfo{person}{Changxing Ding}.} \bibinfo{year}{2024}\natexlab{a}.
\newblock \showarticletitle{Towards variable and coordinated holistic co-speech motion generation}. In \bibinfo{booktitle}{\emph{Proceedings of the IEEE/CVF Conference on Computer Vision and Pattern Recognition}}. \bibinfo{pages}{1566--1576}.
\newblock


\bibitem[Liu et~al\mbox{.}(2025c)]%
        {s2c}
\bibfield{author}{\bibinfo{person}{Zihao Liu}, \bibinfo{person}{Mingwen Ou}, \bibinfo{person}{Zunnan Xu}, \bibinfo{person}{Jiaqi Huang}, \bibinfo{person}{Haonan Han}, \bibinfo{person}{Ronghui Li}, {and} \bibinfo{person}{Xiu Li}.} \bibinfo{year}{2025}\natexlab{c}.
\newblock \showarticletitle{Separate to Collaborate: Dual-Stream Diffusion Model for Coordinated Piano Hand Motion Synthesis}. In \bibinfo{booktitle}{\emph{Proceedings of the 33rd ACM International Conference on Multimedia}}. \bibinfo{pages}{9743--9752}.
\newblock


\bibitem[Mansourian et~al\mbox{.}(2025)]%
        {mansourian2025comprehensive}
\bibfield{author}{\bibinfo{person}{Amir~M Mansourian}, \bibinfo{person}{Rozhan Ahmadi}, \bibinfo{person}{Masoud Ghafouri}, \bibinfo{person}{Amir~Mohammad Babaei}, \bibinfo{person}{Elaheh~Badali Golezani}, \bibinfo{person}{Zeynab~Yasamani Ghamchi}, \bibinfo{person}{Vida Ramezanian}, \bibinfo{person}{Alireza Taherian}, \bibinfo{person}{Kimia Dinashi}, \bibinfo{person}{Amirali Miri}, {et~al\mbox{.}}} \bibinfo{year}{2025}\natexlab{}.
\newblock \showarticletitle{A comprehensive survey on knowledge distillation}.
\newblock \bibinfo{journal}{\emph{arXiv preprint arXiv:2503.12067}} (\bibinfo{year}{2025}).
\newblock


\bibitem[Musy(2018)]%
        {musy_pianoplayer}
\bibfield{author}{\bibinfo{person}{Marco Musy}.} \bibinfo{year}{2018}\natexlab{}.
\newblock \bibinfo{title}{pianoplayer: Automatic fingering generator for piano scores}.
\newblock \bibinfo{howpublished}{\url{https://github.com/marcomusy/pianoplayer}}.
\newblock
\urldef\tempurl%
\url{https://github.com/marcomusy/pianoplayer}
\showURL{%
\tempurl}


\bibitem[Nishizawa et~al\mbox{.}(2025)]%
        {nishizawa2025syncviolinist}
\bibfield{author}{\bibinfo{person}{Hiroki Nishizawa}, \bibinfo{person}{Keitaro Tanaka}, \bibinfo{person}{Asuka Hirata}, \bibinfo{person}{Shugo Yamaguchi}, \bibinfo{person}{Qi Feng}, \bibinfo{person}{Masatoshi Hamanaka}, {and} \bibinfo{person}{Shigeo Morishima}.} \bibinfo{year}{2025}\natexlab{}.
\newblock \showarticletitle{SyncViolinist: Music-Oriented Violin Motion Generation Based on Bowing and Fingering}. In \bibinfo{booktitle}{\emph{2025 IEEE/CVF Winter Conference on Applications of Computer Vision (WACV)}}. IEEE, \bibinfo{pages}{5419--5428}.
\newblock


\bibitem[Peng et~al\mbox{.}(2023)]%
        {peng2023emotalk}
\bibfield{author}{\bibinfo{person}{Ziqiao Peng}, \bibinfo{person}{Haoyu Wu}, \bibinfo{person}{Zhenbo Song}, \bibinfo{person}{Hao Xu}, \bibinfo{person}{Xiangyu Zhu}, \bibinfo{person}{Jun He}, \bibinfo{person}{Hongyan Liu}, {and} \bibinfo{person}{Zhaoxin Fan}.} \bibinfo{year}{2023}\natexlab{}.
\newblock \showarticletitle{Emotalk: Speech-driven emotional disentanglement for 3d face animation}. In \bibinfo{booktitle}{\emph{Proceedings of the IEEE/CVF international conference on computer vision}}. \bibinfo{pages}{20687--20697}.
\newblock


\bibitem[Perez et~al\mbox{.}(2018)]%
        {perez2018film}
\bibfield{author}{\bibinfo{person}{Ethan Perez}, \bibinfo{person}{Florian Strub}, \bibinfo{person}{Harm De~Vries}, \bibinfo{person}{Vincent Dumoulin}, {and} \bibinfo{person}{Aaron Courville}.} \bibinfo{year}{2018}\natexlab{}.
\newblock \showarticletitle{Film: Visual reasoning with a general conditioning layer}. In \bibinfo{booktitle}{\emph{Proceedings of the AAAI conference on artificial intelligence}}, Vol.~\bibinfo{volume}{32}.
\newblock


\bibitem[Richard et~al\mbox{.}(2021)]%
        {richard2021meshtalk}
\bibfield{author}{\bibinfo{person}{Alexander Richard}, \bibinfo{person}{Michael Zollh{\"o}fer}, \bibinfo{person}{Yandong Wen}, \bibinfo{person}{Fernando De~la Torre}, {and} \bibinfo{person}{Yaser Sheikh}.} \bibinfo{year}{2021}\natexlab{}.
\newblock \showarticletitle{Meshtalk: 3d face animation from speech using cross-modality disentanglement}. In \bibinfo{booktitle}{\emph{Proceedings of the IEEE/CVF international conference on computer vision}}. \bibinfo{pages}{1173--1182}.
\newblock


\bibitem[Romero et~al\mbox{.}(2017)]%
        {mano}
\bibfield{author}{\bibinfo{person}{Javier Romero}, \bibinfo{person}{Dimitrios Tzionas}, {and} \bibinfo{person}{Michael~J Black}.} \bibinfo{year}{2017}\natexlab{}.
\newblock \showarticletitle{Embodied hands: Modeling and capturing hands and bodies together}.
\newblock \bibinfo{journal}{\emph{ACM Transactions on Graphics (TOG)}} (\bibinfo{year}{2017}).
\newblock


\bibitem[Ronneberger et~al\mbox{.}(2015)]%
        {ronneberger2015u}
\bibfield{author}{\bibinfo{person}{Olaf Ronneberger}, \bibinfo{person}{Philipp Fischer}, {and} \bibinfo{person}{Thomas Brox}.} \bibinfo{year}{2015}\natexlab{}.
\newblock \showarticletitle{U-net: Convolutional networks for biomedical image segmentation}. In \bibinfo{booktitle}{\emph{International Conference on Medical image computing and computer-assisted intervention}}. Springer, \bibinfo{pages}{234--241}.
\newblock


\bibitem[Rothstein(1995)]%
        {rothstein1995midi}
\bibfield{author}{\bibinfo{person}{Joseph Rothstein}.} \bibinfo{year}{1995}\natexlab{}.
\newblock \bibinfo{booktitle}{\emph{MIDI: A comprehensive introduction}}. Vol.~\bibinfo{volume}{7}.
\newblock \bibinfo{publisher}{AR Editions, Inc.}
\newblock


\bibitem[Shen et~al\mbox{.}(2023)]%
        {shen2023difftalk}
\bibfield{author}{\bibinfo{person}{Shuai Shen}, \bibinfo{person}{Wenliang Zhao}, \bibinfo{person}{Zibin Meng}, \bibinfo{person}{Wanhua Li}, \bibinfo{person}{Zheng Zhu}, \bibinfo{person}{Jie Zhou}, {and} \bibinfo{person}{Jiwen Lu}.} \bibinfo{year}{2023}\natexlab{}.
\newblock \showarticletitle{Difftalk: Crafting diffusion models for generalized audio-driven portraits animation}. In \bibinfo{booktitle}{\emph{Proceedings of the IEEE/CVF conference on computer vision and pattern recognition}}. \bibinfo{pages}{1982--1991}.
\newblock


\bibitem[Siyao et~al\mbox{.}(2022)]%
        {siyao2022bailando}
\bibfield{author}{\bibinfo{person}{Li Siyao}, \bibinfo{person}{Weijiang Yu}, \bibinfo{person}{Tianpei Gu}, \bibinfo{person}{Chunze Lin}, \bibinfo{person}{Quan Wang}, \bibinfo{person}{Chen Qian}, \bibinfo{person}{Chen~Change Loy}, {and} \bibinfo{person}{Ziwei Liu}.} \bibinfo{year}{2022}\natexlab{}.
\newblock \showarticletitle{Bailando: 3d dance generation by actor-critic gpt with choreographic memory}. In \bibinfo{booktitle}{\emph{Proceedings of the IEEE/CVF Conference on Computer Vision and Pattern Recognition}}. \bibinfo{pages}{11050--11059}.
\newblock


\bibitem[Sui et~al\mbox{.}(2026)]%
        {sui2026survey}
\bibfield{author}{\bibinfo{person}{Kewei Sui}, \bibinfo{person}{Anindita Ghosh}, \bibinfo{person}{Inwoo Hwang}, \bibinfo{person}{Bing Zhou}, \bibinfo{person}{Jian Wang}, {and} \bibinfo{person}{Chuan Guo}.} \bibinfo{year}{2026}\natexlab{}.
\newblock \showarticletitle{A survey on human interaction motion generation}.
\newblock \bibinfo{journal}{\emph{International Journal of Computer Vision}} \bibinfo{volume}{134}, \bibinfo{number}{3} (\bibinfo{year}{2026}), \bibinfo{pages}{113}.
\newblock


\bibitem[Sun et~al\mbox{.}(2024)]%
        {sun2024diffposetalk}
\bibfield{author}{\bibinfo{person}{Zhiyao Sun}, \bibinfo{person}{Tian Lv}, \bibinfo{person}{Sheng Ye}, \bibinfo{person}{Matthieu Lin}, \bibinfo{person}{Jenny Sheng}, \bibinfo{person}{Yu-Hui Wen}, \bibinfo{person}{Minjing Yu}, {and} \bibinfo{person}{Yong-jin Liu}.} \bibinfo{year}{2024}\natexlab{}.
\newblock \showarticletitle{Diffposetalk: Speech-driven stylistic 3d facial animation and head pose generation via diffusion models}.
\newblock \bibinfo{journal}{\emph{ACM Transactions on Graphics (ToG)}} \bibinfo{volume}{43}, \bibinfo{number}{4} (\bibinfo{year}{2024}), \bibinfo{pages}{1--9}.
\newblock


\bibitem[Tseng et~al\mbox{.}(2023)]%
        {tseng2023edge}
\bibfield{author}{\bibinfo{person}{Jonathan Tseng}, \bibinfo{person}{Rodrigo Castellon}, {and} \bibinfo{person}{Karen Liu}.} \bibinfo{year}{2023}\natexlab{}.
\newblock \showarticletitle{Edge: Editable dance generation from music}. In \bibinfo{booktitle}{\emph{Proceedings of the IEEE/CVF conference on computer vision and pattern recognition}}. \bibinfo{pages}{448--458}.
\newblock


\bibitem[Wang et~al\mbox{.}(2025b)]%
        {wang2025universe}
\bibfield{author}{\bibinfo{person}{Duomin Wang}, \bibinfo{person}{Wei Zuo}, \bibinfo{person}{Aojie Li}, \bibinfo{person}{Ling-Hao Chen}, \bibinfo{person}{Xinyao Liao}, \bibinfo{person}{Deyu Zhou}, \bibinfo{person}{Zixin Yin}, \bibinfo{person}{Xili Dai}, \bibinfo{person}{Daxin Jiang}, {and} \bibinfo{person}{Gang Yu}.} \bibinfo{year}{2025}\natexlab{b}.
\newblock \showarticletitle{UniVerse-1: Unified Audio-Video Generation via Stitching of Experts}.
\newblock \bibinfo{journal}{\emph{arXiv preprint arXiv:2509.06155}} (\bibinfo{year}{2025}).
\newblock


\bibitem[Wang et~al\mbox{.}(2025a)]%
        {wang2025pamd}
\bibfield{author}{\bibinfo{person}{Hongsong Wang}, \bibinfo{person}{Yin Zhu}, \bibinfo{person}{Qiuxia Lai}, \bibinfo{person}{Yang Zhang}, \bibinfo{person}{Guo-Sen Xie}, {and} \bibinfo{person}{Xin Geng}.} \bibinfo{year}{2025}\natexlab{a}.
\newblock \showarticletitle{PAMD: Plausibility-Aware Motion Diffusion Model for Long Dance Generation}.
\newblock \bibinfo{journal}{\emph{arXiv preprint arXiv:2505.20056}} (\bibinfo{year}{2025}).
\newblock


\bibitem[Wang et~al\mbox{.}(2024)]%
        {wang2024furelise}
\bibfield{author}{\bibinfo{person}{Ruocheng Wang}, \bibinfo{person}{Pei Xu}, \bibinfo{person}{Haochen Shi}, \bibinfo{person}{Elizabeth Schumann}, {and} \bibinfo{person}{C~Karen Liu}.} \bibinfo{year}{2024}\natexlab{}.
\newblock \showarticletitle{F{\"u}relise: Capturing and physically synthesizing hand motion of piano performance}. In \bibinfo{booktitle}{\emph{SIGGRAPH Asia 2024 Conference Papers}}. \bibinfo{pages}{1--11}.
\newblock


\bibitem[Wei et~al\mbox{.}(2024)]%
        {wei2024aniportrait}
\bibfield{author}{\bibinfo{person}{Huawei Wei}, \bibinfo{person}{Zejun Yang}, {and} \bibinfo{person}{Zhisheng Wang}.} \bibinfo{year}{2024}\natexlab{}.
\newblock \showarticletitle{Aniportrait: Audio-driven synthesis of photorealistic portrait animation}.
\newblock \bibinfo{journal}{\emph{arXiv preprint arXiv:2403.17694}} (\bibinfo{year}{2024}).
\newblock


\bibitem[Xu(2025)]%
        {xu2025study}
\bibfield{author}{\bibinfo{person}{Ruipin Xu}.} \bibinfo{year}{2025}\natexlab{}.
\newblock \showarticletitle{Study on teaching and training system construction of piano based on motion capture}. In \bibinfo{booktitle}{\emph{Fourth International Conference on Electronics Technology and Artificial Intelligence (ETAI 2025)}}, Vol.~\bibinfo{volume}{13692}. SPIE, \bibinfo{pages}{718--724}.
\newblock


\bibitem[Xu et~al\mbox{.}(2025)]%
        {xu2025mospa}
\bibfield{author}{\bibinfo{person}{Shuyang Xu}, \bibinfo{person}{Zhiyang Dou}, \bibinfo{person}{Mingyi Shi}, \bibinfo{person}{Liang Pan}, \bibinfo{person}{Leo Ho}, \bibinfo{person}{Jingbo Wang}, \bibinfo{person}{Yuan Liu}, \bibinfo{person}{Cheng Lin}, \bibinfo{person}{Yuexin Ma}, \bibinfo{person}{Wenping Wang}, {et~al\mbox{.}}} \bibinfo{year}{2025}\natexlab{}.
\newblock \showarticletitle{Mospa: Human motion generation driven by spatial audio}.
\newblock \bibinfo{journal}{\emph{arXiv preprint arXiv:2507.11949}} (\bibinfo{year}{2025}).
\newblock


\bibitem[Xu et~al\mbox{.}(2026)]%
        {xu2026emotionally}
\bibfield{author}{\bibinfo{person}{Yifan Xu}, \bibinfo{person}{Sirui Zhao}, \bibinfo{person}{Shifeng Liu}, \bibinfo{person}{Tong Xu}, {and} \bibinfo{person}{Enhong Chen}.} \bibinfo{year}{2026}\natexlab{}.
\newblock \showarticletitle{Emotionally Controllable Audio-driven Talking Face Generation}.
\newblock \bibinfo{journal}{\emph{ACM Transactions on Multimedia Computing, Communications and Applications}} (\bibinfo{year}{2026}).
\newblock


\bibitem[Xu et~al\mbox{.}(2024)]%
        {xu2024mambatalk}
\bibfield{author}{\bibinfo{person}{Zunnan Xu}, \bibinfo{person}{Yukang Lin}, \bibinfo{person}{Haonan Han}, \bibinfo{person}{Sicheng Yang}, \bibinfo{person}{Ronghui Li}, \bibinfo{person}{Yachao Zhang}, {and} \bibinfo{person}{Xiu Li}.} \bibinfo{year}{2024}\natexlab{}.
\newblock \showarticletitle{Mambatalk: Efficient holistic gesture synthesis with selective state space models}.
\newblock \bibinfo{journal}{\emph{Advances in Neural Information Processing Systems}}  \bibinfo{volume}{37} (\bibinfo{year}{2024}), \bibinfo{pages}{20055--20080}.
\newblock


\bibitem[Yang et~al\mbox{.}(2025b)]%
        {yang2025matchdance}
\bibfield{author}{\bibinfo{person}{Kaixing Yang}, \bibinfo{person}{Xulong Tang}, \bibinfo{person}{Yuxuan Hu}, \bibinfo{person}{Jiahao Yang}, \bibinfo{person}{Hongyan Liu}, \bibinfo{person}{Qinnan Zhang}, \bibinfo{person}{Jun He}, {and} \bibinfo{person}{Zhaoxin Fan}.} \bibinfo{year}{2025}\natexlab{b}.
\newblock \showarticletitle{Matchdance: Collaborative mamba-transformer architecture matching for high-quality 3d dance synthesis}.
\newblock \bibinfo{journal}{\emph{arXiv preprint arXiv:2505.14222}} (\bibinfo{year}{2025}).
\newblock


\bibitem[Yang et~al\mbox{.}(2025c)]%
        {yang2025megadance}
\bibfield{author}{\bibinfo{person}{Kaixing Yang}, \bibinfo{person}{Xulong Tang}, \bibinfo{person}{Ziqiao Peng}, \bibinfo{person}{Yuxuan Hu}, \bibinfo{person}{Jun He}, {and} \bibinfo{person}{Hongyan Liu}.} \bibinfo{year}{2025}\natexlab{c}.
\newblock \showarticletitle{Megadance: Mixture-of-experts architecture for genre-aware 3d dance generation}.
\newblock \bibinfo{journal}{\emph{arXiv preprint arXiv:2505.17543}} (\bibinfo{year}{2025}).
\newblock


\bibitem[Yang et~al\mbox{.}(2025a)]%
        {yang2025gesturehydra}
\bibfield{author}{\bibinfo{person}{Quanwei Yang}, \bibinfo{person}{Luying Huang}, \bibinfo{person}{Kaisiyuan Wang}, \bibinfo{person}{Jiazhi Guan}, \bibinfo{person}{Shengyi He}, \bibinfo{person}{Fengguo Li}, \bibinfo{person}{Hang Zhou}, \bibinfo{person}{Lingyun Yu}, \bibinfo{person}{Yingying Li}, \bibinfo{person}{Haocheng Feng}, {et~al\mbox{.}}} \bibinfo{year}{2025}\natexlab{a}.
\newblock \showarticletitle{GestureHYDRA: Semantic Co-speech Gesture Synthesis via Hybrid Modality Diffusion Transformer and Cascaded-Synchronized Retrieval-Augmented Generation}. In \bibinfo{booktitle}{\emph{Proceedings of the IEEE/CVF International Conference on Computer Vision}}. \bibinfo{pages}{12615--12625}.
\newblock


\bibitem[Yang et~al\mbox{.}(2026)]%
        {yang2026streamingtalker}
\bibfield{author}{\bibinfo{person}{Yifan Yang}, \bibinfo{person}{Zhi Cen}, \bibinfo{person}{Sida Peng}, \bibinfo{person}{Xiangwei Chen}, \bibinfo{person}{Yifu Deng}, \bibinfo{person}{Xinyu Zhu}, \bibinfo{person}{Fan Jia}, \bibinfo{person}{Xiaowei Zhou}, {and} \bibinfo{person}{Hujun Bao}.} \bibinfo{year}{2026}\natexlab{}.
\newblock \showarticletitle{StreamingTalker: Audio-driven 3D Facial Animation with Autoregressive Diffusion Model}. In \bibinfo{booktitle}{\emph{Proceedings of the AAAI Conference on Artificial Intelligence}}, Vol.~\bibinfo{volume}{40}. \bibinfo{pages}{11766--11774}.
\newblock


\bibitem[Yi et~al\mbox{.}(2023)]%
        {yi2023generating}
\bibfield{author}{\bibinfo{person}{Hongwei Yi}, \bibinfo{person}{Hualin Liang}, \bibinfo{person}{Yifei Liu}, \bibinfo{person}{Qiong Cao}, \bibinfo{person}{Yandong Wen}, \bibinfo{person}{Timo Bolkart}, \bibinfo{person}{Dacheng Tao}, {and} \bibinfo{person}{Michael~J Black}.} \bibinfo{year}{2023}\natexlab{}.
\newblock \showarticletitle{Generating holistic 3d human motion from speech}. In \bibinfo{booktitle}{\emph{Proceedings of the IEEE/CVF Conference on Computer Vision and Pattern Recognition}}. \bibinfo{pages}{469--480}.
\newblock


\bibitem[Zakka et~al\mbox{.}(2023)]%
        {zakka2023robopianist}
\bibfield{author}{\bibinfo{person}{Kevin Zakka}, \bibinfo{person}{Philipp Wu}, \bibinfo{person}{Laura Smith}, \bibinfo{person}{Nimrod Gileadi}, \bibinfo{person}{Taylor Howell}, \bibinfo{person}{Xue~Bin Peng}, \bibinfo{person}{Sumeet Singh}, \bibinfo{person}{Yuval Tassa}, \bibinfo{person}{Pete Florence}, \bibinfo{person}{Andy Zeng}, {et~al\mbox{.}}} \bibinfo{year}{2023}\natexlab{}.
\newblock \showarticletitle{Robopianist: Dexterous piano playing with deep reinforcement learning}.
\newblock \bibinfo{journal}{\emph{arXiv preprint arXiv:2304.04150}} (\bibinfo{year}{2023}).
\newblock


\bibitem[Zeulner et~al\mbox{.}(2025)]%
        {zeulner2025learning}
\bibfield{author}{\bibinfo{person}{Yves-Simon Zeulner}, \bibinfo{person}{Sandeep Selvaraj}, {and} \bibinfo{person}{Roberto Calandra}.} \bibinfo{year}{2025}\natexlab{}.
\newblock \showarticletitle{Learning to play piano in the real world}.
\newblock \bibinfo{journal}{\emph{arXiv preprint arXiv:2503.15481}} (\bibinfo{year}{2025}).
\newblock


\bibitem[Zhang et~al\mbox{.}(2023)]%
        {zhang2023sadtalker}
\bibfield{author}{\bibinfo{person}{Wenxuan Zhang}, \bibinfo{person}{Xiaodong Cun}, \bibinfo{person}{Xuan Wang}, \bibinfo{person}{Yong Zhang}, \bibinfo{person}{Xi Shen}, \bibinfo{person}{Yu Guo}, \bibinfo{person}{Ying Shan}, {and} \bibinfo{person}{Fei Wang}.} \bibinfo{year}{2023}\natexlab{}.
\newblock \showarticletitle{Sadtalker: Learning realistic 3d motion coefficients for stylized audio-driven single image talking face animation}. In \bibinfo{booktitle}{\emph{Proceedings of the IEEE/CVF conference on computer vision and pattern recognition}}. \bibinfo{pages}{8652--8661}.
\newblock


\bibitem[Zhi et~al\mbox{.}(2023)]%
        {zhi2023livelyspeaker}
\bibfield{author}{\bibinfo{person}{Yihao Zhi}, \bibinfo{person}{Xiaodong Cun}, \bibinfo{person}{Xuelin Chen}, \bibinfo{person}{Xi Shen}, \bibinfo{person}{Wen Guo}, \bibinfo{person}{Shaoli Huang}, {and} \bibinfo{person}{Shenghua Gao}.} \bibinfo{year}{2023}\natexlab{}.
\newblock \showarticletitle{Livelyspeaker: Towards semantic-aware co-speech gesture generation}. In \bibinfo{booktitle}{\emph{Proceedings of the IEEE/CVF international conference on computer vision}}. \bibinfo{pages}{20807--20817}.
\newblock


\bibitem[Zhu et~al\mbox{.}(2025)]%
        {zhu2025muq}
\bibfield{author}{\bibinfo{person}{Haina Zhu}, \bibinfo{person}{Yizhi Zhou}, \bibinfo{person}{Hangting Chen}, \bibinfo{person}{Jianwei Yu}, \bibinfo{person}{Ziyang Ma}, \bibinfo{person}{Rongzhi Gu}, \bibinfo{person}{Yi Luo}, \bibinfo{person}{Wei Tan}, {and} \bibinfo{person}{Xie Chen}.} \bibinfo{year}{2025}\natexlab{}.
\newblock \showarticletitle{Mu{Q}: Self-supervised music representation learning with mel residual vector quantization}.
\newblock \bibinfo{journal}{\emph{IEEE Transactions on Audio, Speech and Language Processing}} (\bibinfo{year}{2025}).
\newblock


\bibitem[Zhu et~al\mbox{.}(2023)]%
        {zhu2023human}
\bibfield{author}{\bibinfo{person}{Wentao Zhu}, \bibinfo{person}{Xiaoxuan Ma}, \bibinfo{person}{Dongwoo Ro}, \bibinfo{person}{Hai Ci}, \bibinfo{person}{Jinlu Zhang}, \bibinfo{person}{Jiaxin Shi}, \bibinfo{person}{Feng Gao}, \bibinfo{person}{Qi Tian}, {and} \bibinfo{person}{Yizhou Wang}.} \bibinfo{year}{2023}\natexlab{}.
\newblock \showarticletitle{Human motion generation: A survey}.
\newblock \bibinfo{journal}{\emph{IEEE Transactions on Pattern Analysis and Machine Intelligence}} \bibinfo{volume}{46}, \bibinfo{number}{4} (\bibinfo{year}{2023}), \bibinfo{pages}{2430--2449}.
\newblock


\end{thebibliography}

\appendix

\end{document}